\documentclass[runningheads]{llncs}

\usepackage{eccv}

\usepackage{eccvabbrv}

\usepackage{graphicx}
\usepackage{booktabs}
\usepackage{authblk}

\newcommand{\equalcontrib}{\textsuperscript{*}}
\newcommand{\crosssymbol}{\textsuperscript{\dag}}
\newcommand{\ddagsymbol}{\textsuperscript{\ddag}}

\usepackage[accsupp]{axessibility}  

\usepackage{hyperref}

\usepackage{orcidlink}

\usepackage{colortbl}
\usepackage{adjustbox}
\usepackage{multirow}
\usepackage{subcaption}
\usepackage{float}
\usepackage{placeins}

\usepackage{algorithm}
\usepackage{algcompatible}
\newcommand{\RETURN}{\State \textbf{return: }}

\usepackage[dvipsnames]{xcolor}

\newcommand{\mycomment}[1]{}

\begin{document}

\title{Multimodal Label Relevance Ranking via Reinforcement Learning} 


\author{Taian Guo\inst{1}\equalcontrib\orcidlink{0000-0003-0787-511X} \and
Taolin Zhang\inst{2}\equalcontrib\crosssymbol\orcidlink{0009-0006-2441-2861} 
\and
Haoqian Wu\inst{1}\orcidlink{0000-0003-1035-1499} \and
Hanjun Li\inst{1}\orcidlink{0009-0006-4211-7479} \and
Ruizhi Qiao\inst{1}\ddagsymbol\orcidlink{0000-0002-3663-0149} \and
Xing Sun\inst{1}\orcidlink{0000-0001-8132-9083}
}
%
\authorrunning{T. Guo et al.}
%

\institute{
Tencent Youtu Lab \\
\email{\{taianguo,linuswu,hanjunli,ruizhiqiao,winfredsun\}@tencent.com}
\and
Tsinghua Shenzhen International Graduate School, Tsinghua University\\
\email{ztl23@mails.tsinghua.edu.cn}
}
%
\let\oldthefootnote\thefootnote
\renewcommand{\thefootnote}{\textsuperscript{*}}
\footnotetext{Equal contribution.}
\renewcommand{\thefootnote}{\textsuperscript{\dag}}
\footnotetext{Work done during internship at Tencent.}
\renewcommand{\thefootnote}{\textsuperscript{\ddag}}
\footnotetext{Corresponding author: Ruizhi Qiao.}
\let\thefootnote\oldthefootnote


\maketitle

\begin{abstract}
Conventional multi-label recognition methods often focus on label confidence, frequently overlooking the pivotal role of partial order relations consistent with human preference. To resolve these issues, we introduce a novel method for multimodal label relevance ranking, named Label Relevance Ranking with Proximal Policy Optimization (LR\textsuperscript{2}PPO), which effectively discerns partial order relations among labels. LR\textsuperscript{2}PPO first utilizes partial order pairs in the target domain to train a reward model, which aims to capture human preference intrinsic to the specific scenario. Furthermore, we meticulously design state representation and a policy loss tailored for ranking tasks, enabling LR\textsuperscript{2}PPO to boost the performance of label relevance ranking model and largely reduce the requirement of partial order annotation for transferring to new scenes. To assist in the evaluation of our approach and similar methods, we further propose a novel benchmark dataset, LRMovieNet, featuring multimodal labels and their corresponding partial order data. Extensive experiments demonstrate that our LR\textsuperscript{2}PPO algorithm achieves state-of-the-art performance, proving its effectiveness in addressing the multimodal label relevance ranking problem. Codes and the proposed LRMovieNet dataset are publicly available at \url{https://github.com/ChazzyGordon/LR2PPO}.
\end{abstract}

\section{Introduction}
\label{sec:intro}

\begin{figure}[t]
\centering
\includegraphics[width=0.8\textwidth, page=1]{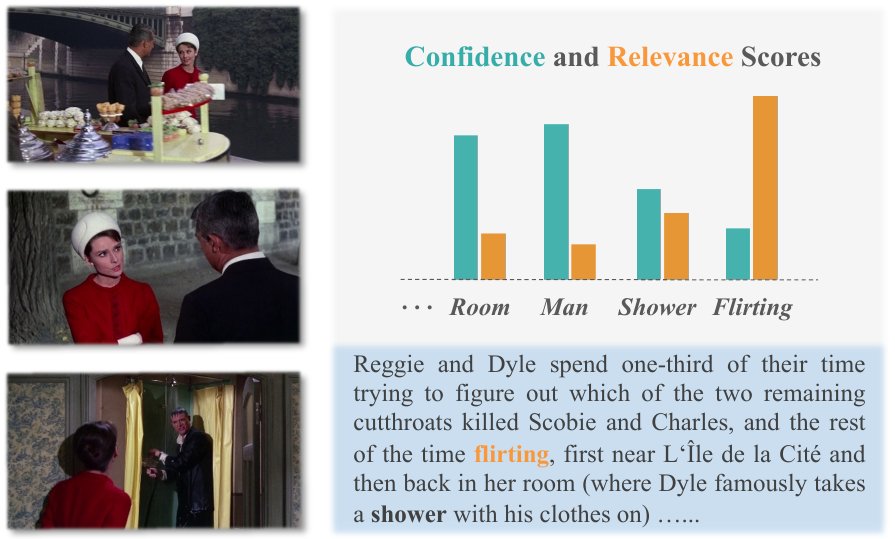}
\caption{
\textbf{
Illustration of the Difference between Label Confidence and Label Relevance.}
This figure provides an example of a movie footage consisting of three consecutive keyframes and its scene description. Generally, conventional label confidence tends to place more emphasis on the tangible objects, whereas the proposed label relevance better reveals the relations between labels and the real scene which they correspond to. As shown in the top right histogram, label confidence models tend to assign a higher level of confidence to the label `\textit{Man}' due to its higher frequency of occurrence within the context. In contrast, the label `\textit{Flirting}' is more closely aligned with the primary theme of the movie scene, 
resulting in a higher label relevance score.
}
\label{fig:teaser}
\end{figure}

Multi-label recognition, a fundamental task in computer vision, aims to identify all possible labels contained within a variety of media forms such as images and videos.
Visual or multimodal recognition is broadly applied in areas such as scene understanding, intelligent content moderation, recommendation systems, surveillance systems, and autonomous driving.
However, due to the complexity of the real world, simple label recognition often proves to be insufficient, as it treats all predicted labels equally without considering the priority of human preferences.
A feasible solution could be to rank all the labels according to their relevance to a specific scene. This approach would allow participants to focus on labels with high scene relevance, while reducing the importance of secondary labels, regardless of their potentially high label confidence.

In contrast to predicting label confidence, the task of label relevance constitutes a more challenging problem that cannot be sufficiently tackled via methods such as calibration \cite{kumar2019verified, li2020learning, garg2020unified, liu2022data}, as the main objective of these methods is to correct label biases or inaccuracies to improve model performance, rather than establishing relevance between the label and the input data.
As illustrated in Fig.~\ref{fig:teaser},
label confidence typically refers to the estimation from a model about the probability of a label's occurrence, while label relevance primarily denotes the significance of the label to the primary theme of multimodal inputs. 
The observation also demonstrates that relevance labels bear a closer alignment with human preferences.
Understandably, ranking the labels in order of relevance can be employed to emphasize the important labels.

Recently, Learning to Rank (LTR) methods \cite{crammer2001pranking, cossock2006subset, li2007mcrank, cao2007learning, xia2008listwise, liu2009learning, burges2010ranknet, ai2018learning, zhuang2018globally, pang2020setrank, buyl2023rankformer} have been explored to tackle ranking problems. However, the primary focus of ranking techniques is based on retrieved documents or recommendation lists, rather than on the target label set. 
Therefore, these approaches are not directly and effectively applicable to address the problem of label relevance ranking due to the data and task setting.

In addition to LTR, there is a more theoretical branch of research that deals with the problem of ranking labels, known as label ranking \cite{har2002constraint, dekel2003log, brinker2006case, shalev2006efficient, brinker2007case, hullermeier2008label, cheng2008instance, furnkranz2008multilabel, cheng2009decision, kanehira2016multi, korba2018structured, dery2020improving, fotakis2022linear, fotakis2022label}. 
However, previous works on label ranking are oblivious to the semantic information of the label classes. 
Therefore, the basis for ranking is the difference between the positive and negative instances within each class label, rather than truly ranking the labels based on the difference in relevance between the label and the input.

Recognizing the significance of label relevance and acknowledging the research gap in this field, our study pioneers an investigation into the relevance between labels and multimodal inputs of video clips. 
We rank the labels with the relevance scores, thereby facilitating participants to extract primary labels of the clip.
To show the difference in relevance between distinct labels and the video clips, we develop a multimodal label relevance dataset LRMovieNet with relevance categories annotation. 
Expanded from MovieNet \cite{huang2020movienet}, LRMovieNet contains various types of multimodal labels and a broad spectrum of label semantic levels, making it more capable of representing situations in the real world.

Intuitively, label relevance ranking can be addressed by performing a simple regression towards the ground truth relevance score.
However, this approach has some obvious shortcomings: 
firstly, the definition of relevance categories does not perfectly conform to human preferences for label relevance.
Secondly, the range of relevance scores cannot accurately distinguish the differences in relevance between labels, which limits the accuracy of the label relevance ranking model, especially when transferring to a new scenario.
Given the above shortcomings, it is necessary to design a method that directly takes advantage of the differences in relevance between different labels and multimodal inputs to better and more efficiently 
transfer the label relevance ranking ability from an existing scenario to a new one.
For clarity, we term the original scenario as source domain, and the new scenario with new labels or new video clips as target domain.
 
By introducing a new state definition and policy loss suitable for the label relevance ranking task, our LR\textsuperscript{2}PPO algorithm is able to effectively utilize the partial order relations in the target domain. 
This makes the ranking model more in line with human preferences, significantly improving the performance of the label relevance ranking algorithm.
Specifically, we train a reward model over the partial order annotation to align with human preference in the target domain, 
and then utilize it to guide the training of the LR\textsuperscript{2}PPO framework.
It is sufficient to train the reward model using a few partial order annotations from the target domain, along with partial order pair samples augmented from the source domain.
Since partial order annotations can better reflect human preferences for primary labels compared to relevance category definitions, this approach can effectively improve the label relevance ranking performance in the target domain.

The main contributions of our work can be summarized as follows:

\begin{enumerate}
\item 
We recognize the significant role of label relevance,
and analyze the limitations of previous ranking methods when dealing with label relevance.
To solve this problem, we propose a multimodal label relevance ranking approach to rank the labels according to the relevance between label and the multimodal input. 
To the best of our knowledge, this is the first work to explore 
the ranking in the perspective of label relevance.

\item 
To better generalize the capability to new scenarios, 
we design a paradigm that transfers 
label relevance ranking ability from the source domain to the target domain. 
Besides, we propose the LR\textsuperscript{2}PPO (Label Relevance Ranking with Proximal Policy Optimization) to effectively mine the partial order relations among labels.

\item 
To better evaluate the effectiveness of LR\textsuperscript{2}PPO, 
we annotate each video clip with corresponding class labels and their relevance order of the MovieNet dataset \cite{huang2020movienet}, and develop a new multimodal label relevance ranking benchmark dataset, LRMovieNet (Label Relevance of MovieNet). 
Comprehensive experiments on this dataset and traditional LTR datasets demonstrate the effectiveness of our proposed LR\textsuperscript{2}PPO algorithm.
\end{enumerate}

\section{Related Works}
\label{sec:related}

\subsection{Learning to Rank}
Learning to rank methods can be categorized into pointwise, pairwise, and listwise approaches. Classic algorithms include Subset Ranking \cite{cossock2006subset}, McRank \cite{li2007mcrank}, Prank \cite{crammer2001pranking} (pointwise), RankNet, LambdaRank, LambdaMart \cite{burges2010ranknet} (pairwise), and ListNet \cite{cao2007learning}, ListMLE \cite{xia2008listwise}, DLCM \cite{ai2018learning}, SetRank \cite{pang2020setrank}, RankFormer \cite{buyl2023rankformer} (listwise). Generative models like miRNN \cite{zhuang2018globally} estimate the entire sequence directly for optimal sequence selection. These methods are primarily used in information retrieval and recommender systems, and differ from label relevance ranking in that they typically rank retrieved documents or recommendation list rather than labels.
Meanwhile, label ranking \cite{har2002constraint, dekel2003log, brinker2006case, shalev2006efficient, brinker2007case, hullermeier2008label, cheng2008instance, furnkranz2008multilabel, cheng2009decision, kanehira2016multi, korba2018structured, dery2020improving, fotakis2022linear, fotakis2022label} is a rather theoretical research field that investigates the relative order of labels in a closed label set.
These methods typically lack perception of the textual semantic information of categories, mainly learning the order relationship based on the difference between positive and negative instances in the training set, rather than truly according to the relevance between the label and the input. 
In addition, these methods heavily rely on manual annotation, which also limits their application in real-world scenarios.
Moreover, these methods have primarily focused on single modality, mainly images, and object labels, suitable for relatively simple scenarios. 
These methods differ significantly from our proposed LR\textsuperscript{2}PPO, which for the first time explores the ranking of multimodal labels according to the relevance between labels and input.
LR\textsuperscript{2}PPO also handles a diverse set of multimodal labels, including not only objects but also events, attributes, and character identities, which are often more challenging and crucial in real-world multimodal video label relevance ranking scenarios.

\subsection{Reinforcement Learning}
Reinforcement learning is a research field of great significance.
Classic reinforcement learning methods, including algorithms like Monte Carlo \cite{barto1993monte}, Q-Learning \cite{watkins1989learning}, DQN \cite{mnih2013playing}, DPG \cite{silver2014deterministic}, DDPG \cite{lillicrap2015continuous}, TRPO \cite{schulman2015trust}, \etc, are broadly employed in gaming, robot control, financial trading, \etc 
Recently, Proximal Policy Optimization (PPO \cite{schulman2017proximal}) algorithm proposed by OpenAI enhances the policy update process, achieving significant improvements in many tasks. 
InstructGPT \cite{ouyang2022training} adopts PPO for human preference feedback learning, significantly improving the performance of language generation. 
In order to make the ranking model more effectively understand the human preference inherent in the partial order annotation from the target domain, we adapt the Proximal Policy Optimization (PPO) algorithm to the label relevance ranking task.
By designing state definitions and policy loss tailored for label relevance ranking, partial order relations are effectively mined in accordance with human preference, improving the performance of label relevance ranking model.

\subsection{Vision-Language Pretraining}
Works in this area include two-stream models like CLIP \cite{clip}, ALIGN \cite{jia2021scaling}, and single-stream models like ViLBERT \cite{lu2019vilbert}, UNITER \cite{chen2020uniter}, UNIMO \cite{li2020unimo}, SOHO \cite{huang2021seeing}, ALBEF \cite{li2021align}, VLMO \cite{bao2022vlmo}, TCL \cite{yang2022vision}, X-VLM \cite{zeng2021multi}, BLIP \cite{li2022blip}, BLIP2 \cite{li2023blip}, CoCa \cite{yu2022coca}.
These works are mainly applied in tasks like Visual Question Answering (VQA), visual entailment, visual grounding, multimodal retrieval, \etc
Our proposed LR\textsuperscript{2}PPO also applies to multimodal inputs, using two-stream Transformers to extract visual and textual features, which are then fused through the cross attention module for subsequent label relevance score prediction.

\section{Method}
\label{sec:method}

\subsection{Preliminary}


\begin{definition}[Label Confidence]
Given a multi-label classification task with a set of labels $\mathcal{L} = \{l_1, l_2, \ldots, l_n\}$, an instance $x$ is associated with a label subset $\mathcal{L}_x \subseteq \mathcal{L}$. The label confidence of a label $l_i$ for instance $x$, denoted as $C(l_i|x)$, is defined as the \textbf{probability} that $l_i$ is a \textbf{correct} label for $x$, \ie,
\begin{equation}
C(l_i|x) = P(l_i \in \mathcal{L}_x|x).
\end{equation}
\end{definition}

\begin{definition}[Label Relevance]
The label relevance of a label $l_i$ for instance $x$, denoted as $R(l_i|x)$, is defined as the \textbf{degree} of \textbf{association} between $l_i$ and $x$, \ie,
\begin{equation}
R(l_i|x) = f(l_i, x),
\end{equation}
where $f$ is a function that measures the degree of association between $l_i$ and $x$.
\end{definition}

\noindent
Given $V$ video clips, where the $j$-th clip consists of frames ${F}^j = [F^j_0, F^j_1, ..., F^j_{N-1}]$, with $N$ representing the total number of frames extracted from a video clip, 
and $j$ ranging from 0 to $V-1$. Each video clip is accompanied by text descriptions ${T}^j$ and a set of recognized labels denoted as $\mathcal{L}^j$, where $\mathcal{L}^j=\{l^j_0, l^j_1,...,l^j_i,...,l^j_{|\mathcal{L}^j|-1}\}$, and $|\mathcal{L}^j|$ is the number of labels in the $j$-th video clip.
The objective of label relevance ranking is to learn a ranking function $f_{\text{rank}}: {F}^j, {T}^j, \mathcal{L}^j \rightarrow {U}^j$, where ${U}^j=[u^j_0, u^j_1,..., u^j_i,...,u^j_{|\mathcal{L}^j|-1}]$ represents the ranking result of the label set $\mathcal{L}^j$.


\subsection{Label Relevance Ranking with Proximal Policy Optimization (LR\textsuperscript{2}PPO)}
\label{subsec:LR2PPO}
\begin{figure*}[ht]
    \centering
    \includegraphics[width=0.83\textwidth, page=1]{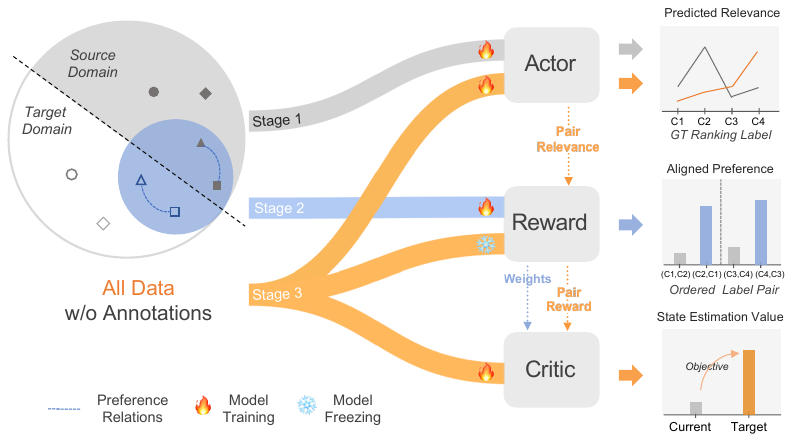}
    \caption{\textbf{Illustration of the training paradigm of LR\textsuperscript{2}PPO}. Each stage takes multimodal data as input but differs in terms of specific data division and annotation type. Technically, in \textcolor[RGB]{140,140,140}{Stage 1}, data from the source domain is employed to establish a label relevance ranking base model (\ie, \textit{\textbf{Actor}}). \textcolor[RGB]{158,182,244}{Stage 2} involves preference data to train a \textit{\textbf{Reward}} model. Finally, in \textcolor[RGB]{239,134,0}{Stage 3}, \textit{\textbf{Critic}} model interacts with the first two models and all data w/o annotations is utilized to boost the performance of the \textbf{\textit{Actor}}, which will solely be applied in the inference stage.} 
    \label{fig:lrppo}
\end{figure*}

We now present the details of our LR\textsuperscript{2}PPO.
As depicted in Fig.~\ref{fig:lrppo}, LR\textsuperscript{2}PPO primarily consists of three network modules: actor, reward, and critic, and the training can be divided into 3 stages. 
We first discuss the relations between three training stages, and then we provide a detailed explanation of each stage.

\noindent
\textbf{Relations between 3 Training Stages.}
We use a two-stream Transformer to handle multimodal input and train three models in LR\textsuperscript{2}PPO: actor, reward, and critic. 
In stage 1, the actor model is supervised on the source domain to obtain a label relevance ranking base model. Stages 2 and 3 generalize the actor model to the target domain.  
In stage 2, the reward model is trained with a few annotated label pairs in the target domain and augmented pairs in the source domain.
In stage 3, reward and critic networks are initialized with the stage 2 reward model.
Then, actor and critic are jointly optimized under LR\textsuperscript{2}PPO with label pair data, guided by the stage 2 reward model, to instruct the actor network with partial order relations in the target domain.
Finally, the optimized actor network, structurally identical to the stage 1 actor, is used as the final label relevance ranking network, adding no inference overhead.

\noindent
\textbf{Stage 1. Label Relevance Ranking Base Model.} 
During Stage 1, the training of the label relevance ranking base model adopts a supervised paradigm, \ie, it is trained on the source domain based on manually annotated relevance categories (high, medium and low).
The label relevance ranking base network accepts multimodal inputs of multiple video frames $F^j$, text descriptions $T^j$, and text label set $\mathcal{L}^j$.
For a video clip, each text label is concatenated with text descriptions.
Then ViT \cite{vit} and Roberta \cite{liu2019roberta} are utilized to extract visual and textual features, respectively.
Subsequently, the multimodal features are fused through the cross attention module.
Finally, through the regression head, the relevance score of each label is predicted, and the SmoothL1Loss 
is calculated based on the manually annotated relevance categories as the final loss, which can be formulated as:
\begin{equation}
\label{eq:smoothl1_loss}
L_{\text{SmoothL1}}(p) = 
\begin{cases}
  0.5 (p - y)^2 / \beta & \text{if } |p - y| < \beta \\
  |p - y| - 0.5 \beta & \text{otherwise},
\end{cases}
\end{equation}
where $p$ and $y$ are the predicted and ground truth relevance score of a label in a video clip, respectively. 
$y = 2,1,0$ corresponds to  high, medium and low relevance, respectively. $\beta$ is a hyper-parameter.


\noindent
\textbf{Stage 2. Reward Model.} 
We train a reward model on the target domain in stage 2. 
With a few label pair annotations on the target domain, along with augmented pairs sampled from the source domain, the reward model can be trained to assign rewards to the partial order relationships between label pairs of a given clip.
This kind of partial order relation annotation aligns with human perference of label relevance ranking, thus benefiting relevance ranking performance with limited annotation data.
In a video clip, the reward model takes the concatenation of the initial pair of labels and the same labels in reranked order (in ground truth order of relevance or in reverse) as input, and predicts the reward of the pair in reranked order through the initial pair.
The loss function adopted for the training of the reward model can be formulated as:
\begin{equation}
\label{eq:reward_model_loss}
\begin{split}
L_{RM}(g_{ini}, g_c) = \max(0, m_R - (R([g_{ini}, g_c]) - R([g_{ini}, \operatorname{flip}(g_c)]))),
\end{split}
\end{equation}
where $g_{ini}$ and $g_c$ represent the initial label pair and the pair in ground truth order respectively, 
$[\cdot, \cdot]$ means concatenation of two label pairs, $\operatorname{flip}(\cdot)$ means flipping the order of the label pair, and $R(\cdot)$ denotes the reward model. $m_R$ is a margin hyperparameter.

\noindent
\textbf{Stage 3. LR\textsuperscript{2}PPO.} 
Stage 3 builds upon the first two training stages to jointly train the LR\textsuperscript{2}PPO framework.
To better address the issues in the label relevance ranking, we modify the state definition and policy loss in the original PPO. 
We redefine the state $s_t$ as the order of a group of labels (specifically, a label pair) at timestep $t$, with the initial state $s_0$ being the original label order at input.
The policy network (aka. actor model) predicts the relevance score of the labels and ranks them from high to low to obtain a new label order as next state $s_{t+1}$, which is considered a state transition, or action $a_t$. 
This process can be regarded as the policy $\pi_\theta$ of state transition. 
The combination of state $s_t$ (the initial label pair) and action $a_t$ (implicitly representing the reranked pair) is evaluated by the reward model to obtain a reward $r_t$.


We denote the target value function estimate at time step $t$ as  $V^{target}_t$, 
which can be formulated as:
\begin{align} 
\label{eq:v_t_target}
V_t^{\text {target }}=r_t+\gamma r_{t+1}+\gamma^2 r_{t+2}+\ldots+\gamma^{T-t-1} r_{T-1} 
+\gamma^{T-t} V_{\omega_{old}}\left(s_T\right),
\end{align}
where
$V_{\omega_{old}}(s_T)$ denotes the old value function estimate at state $s_T$. 
$\omega$ is the trainable parameters of an employed state value network (aka. critic model)
$V_{\omega}(\cdot)$,
while $V_{\omega_{old}}(\cdot)$ represents the old state value network.
$\gamma$ is the discount factor.
$T$ is the terminal time step.
We can further obtain the advantage  $\hat{A}_t$ estimate at time step $t$ via the target value $V_t^{target}$ and the critic model's prediction for the value of the current state $s_t$:
\begin{equation}
\label{eq:advantage}
    \hat{A}_t = V_t^{target} - V_{\omega_{old}}(s_t),
\end{equation}
where 
$V_{\omega_{old}}(s_t)$ denotes the old value function estimate at state $s_t$. 

In typical reinforcement learning tasks, such as gaming, decision control, language generation, \etc, 
PPO usually takes the maximum component in the vector of predicted action probability distribution to obtain the ratio item of policy loss. 
(See supplementary for more details.)
However, in the label relevance ranking task, it requires a complete probability vector to represent the change in label order, \ie, a state transition. Thus, it is difficult to directly build the ratio item of policy loss from action probability.

To address this problem, 
we first define the partial order function, \ie:
\begin{equation}
\label{eq:partial_ordering_loss}
H_{partial}(p_t^1,p_t^2)=\operatorname{max}(0,m-(p_t^1-p_t^2)), 
\end{equation}
where $p_t^1$ and $p_t^2$ 
represent the predicted scores of the input label pair by the actor network at time step $t$ 
, \ie, $\pi_{\theta}(a_t|s_t)$
, and $m$ is a margin hyperparameter, which helps the model better distinguish between correct and incorrect predictions to make the policy loss more precise. 

Maximizing the surrogate objective is equal to minimize the policy loss in the context of reinforcement learning.
In original PPO, ratio $r_t(\theta)$ is adopted to measure the change in policy 
and serves as a multiplication factor in the surrogate objective to
encourage the policy to increase the probability of actions that have a positive advantage, and decrease the probability of actions that have a negative advantage.
However, in label relevance ranking, determining the action for a single state transition necessitates a comprehensive label sequence probability distribution. 
Consequently, if we employ the ratio calculation approach from the original PPO, the ratio fails to encapsulate the change between the new and old policies, thereby inhibiting effective adjustment of the advantage within the surrogate objective. 
This issue persists even when adopting the clipped objective function.
Please refer to Sec.~\ref{sec:ablation} for more experimental analysis.
To solve this problem, we propose partial order ratio $r_t^{\prime}(\theta)$ to provide a more suitable adjustment for advantage in the surrogate objective.
It is a function that depends on the sign of $\hat{A}_t$, the estimated advantage at time step $t$. 
In practice, we utilize a small negative threshold $\delta$ instead of zero to stabilize the joint training of LR\textsuperscript{2}PPO framework.
Specifically, $r_t^{\prime}(\theta)$ is formulated as:
\begin{equation}
\label{eq:r_t_prime}
r_t^{\prime}(\theta) = \begin{cases} -H^{partial}(p_t^1,p_t^2) & \hat{A}_t \geq \delta \\     -H^{partial}(p_t^2,p_t^1) & \hat{A}_t < \delta. \end{cases}
\end{equation}
The proposed partial order ratio $r_t^{\prime}(\theta)$ encourages the model to correctly rank the labels by penalizing incorrect orderings.
In Eq.~(\ref{eq:r_t_prime}), assuming the advantage $\hat{A}_t$ surpasses $\delta$ (\ie, $\hat{A}_t \geq \delta$), the reward model favors the first label (scored $p_t^1$) over the second (scored $p_t^2$). If $p_t^1 > p_t^2$, the absolute value of $r_t^{\prime}(\theta)$ falls below $m$, lessening the penalty in Eq.~(\ref{eq:lrppo_policy_loss}). Conversely, if the advantage is below $\delta$, the opposite happens.
The policy function loss of LR\textsuperscript{2}PPO is formulated as:
\begin{equation} \label{eq:lrppo_policy_loss} 
L_{LR\textsuperscript{2}PPO}^{PF}(\theta) = - \mathbb{E}_t \left( r^{\prime}_{t}(\theta) abs(\hat{A}_t) \right).
\end{equation}
In our design, the order of label pairs is adjusted based on the relative magnitude of the advantage $\hat{A}_t$ and $\delta$.
The absolute value function $abs(\cdot)$ in Eq.~(\ref{eq:lrppo_policy_loss}) ensures that the advantage is always positive, reflecting the fact that moving a more important label to a higher position is beneficial.
At the same time, the advantage is taken as an absolute value, indicating that after adjusting the order of the labels, \ie, moving the more important label to the front 
(according to the relative magnitude of the original advantage $\hat{A}_t$ and $\delta$), 
the advantage can maintain a positive value. 
In this way, policy loss can be more suitable for the label relevance ranking task, ensuring the ranking performance of LR\textsuperscript{2}PPO.

Meanwhile, as original PPO, the value function loss of LR\textsuperscript{2}PPO is given by:
\begin{equation} 
\label{eq:lrppo_value_loss}
L_{LR\textsuperscript{2}PPO}^{VF} (\omega) = L^{VF}(\omega) = \mathbb{E}_t \left[ \left( V_\omega(s_t) - V^{target}_t \right)^2 \right].
\end{equation}
This is the expected value at time step $t$ of the squared difference between the value function estimate $V_\omega(s_t)$ and the target value function estimate $V^{target}_t$, under the policy parameters $\theta$.
This loss function measures the discrepancy between the predicted and actual value functions, driving the model to better estimate the expected return.


As original PPO, we utilize entropy bonus $S(\pi_\theta)$ to encourage exploration by maximizing the entropy of the policy, and employ KL penalty $KL_{penalty} (\pi_{\theta_{old}}, \pi_{\theta}) $ to constrain policy updates and to prevent large performance drops during optimization.
Please refer to supplementary for more details.
Finally, the overall loss function of LR\textsuperscript{2}PPO combines the policy function loss, the value function loss, the entropy bonus, and the KL penalty for encouraging exploration:
\begin{align} 
\label{eq:lrppo_loss}
L_{LR\textsuperscript{2}PPO}(\theta,\omega) = &L_{LR\textsuperscript{2}PPO}^{PF}(\theta) + c_1 L_{LR\textsuperscript{2}PPO}^{VF}(\omega) \nonumber \\
&- c_2 S(\pi_\theta) + c_3 KL_{penalty} (\pi_{\theta_{old}}, \pi_{\theta}),
\end{align}
where $c_1$, $c_2$ and $c_3$ are hyper-parameter coefficients.

In summary, our LR\textsuperscript{2}PPO algorithm leverages a combination of a label relevance ranking base model, a reward model, and a critic model, trained in a three-stage process. 
The algorithm is guided by a carefully designed loss function that encourages correct label relevance ranking, accurate value estimation, sufficient exploration, and imposes a constraint on policy updates.
The pseudo-code of our LR\textsuperscript{2}PPO is provided in Algorithm \ref{alg:lrppo}.

\begin{algorithm}
\caption{Label Relevance Ranking with Proximal Policy Optimization (LR\textsuperscript{2}PPO), Actor-Critic Style}
\label{alg:lrppo}
\begin{algorithmic}[1]
\renewcommand{\algorithmicrequire}{\textbf{Input:}}
\renewcommand{\algorithmicensure}{\textbf{Output:}}
\REQUIRE Policy network $\pi_{\theta_{\text{old}}}$, state value network $V_{\omega_{\text{old}}}$, number of timesteps $T$, 
number of trajectories in an iteration $N_{\text{Trajs}}$, 
number of epochs $K$, minibatch size $M$
\ENSURE Policy network parameter $\theta$, state value network parameter $\omega$
\\ \textit{Initialization}:
\STATE Initialize $\theta_{\text{old}}$ and $\omega_{\text{old}}$ with base model and reward model
\\ \textit{LOOP Process}
\FOR {iteration = 1, 2, ...}
    \FOR{$n_{\text{traj}}=1,2,\ldots,{N_{\text{Trajs}}}$}
        \STATE Run policy $\pi_{\theta_{\text{old}}}$ and state value network $V_{\omega_{\text{old}}}$ in environment for $T$ timesteps
        \STATE Compute advantage estimates $\hat{A}_1, \ldots, \hat{A}_T$ according to Eq.~(\ref{eq:advantage})
    \ENDFOR
    \STATE Compute joint loss $L_{\text{LR}^2\text{PPO}}$ according to Eq.~(\ref{eq:lrppo_loss})
    \STATE Optimize surrogate $L_{\text{LR}^2\text{PPO}}$ with respect to $\theta$ and $\omega$, with $K$ epochs and minibatch size $M \leq N_{\text{Trajs}} \cdot T$
    \STATE $\theta_{\text{old}} \leftarrow \theta$, $\omega_{\text{old}} \leftarrow \omega$
\ENDFOR
\RETURN $\theta$, $\omega$
\end{algorithmic} 
\end{algorithm}

\section{Label Relevance Ranking Dataset}
\label{sec:dataset}

Our objective in this paper is to tackle the challenge of label relevance ranking within multimodal scenarios, with the aim of better identifying salient and core labels in these contexts. 
However, we find that existing label ranking datasets are often designed with a focus on single-image inputs, lack text modality, and their fixed label systems limit the richness and diversity of labels. 
Furthermore, datasets related to Learning to Rank are typically tailored for tasks such as document ranking, making them unsuitable for label relevance ranking tasks.

Our proposed method for multimodal label relevance ranking is primarily designed for multimodal scenarios that feature a rich and diverse array of labels, particularly in typical scenarios like video clips. In our search for suitable datasets, we identify  the MovieNet dataset as a rich source of multimodal video data. However, the MovieNet dataset only provides image-level object label annotations, while the wealth of information available in movie video clips can be used to extract a broad range of multimodal labels.
To address this gap, we have undertaken a process of further label extraction and cleaning from the MovieNet dataset, with the aim of transforming the benchmark for label relevance ranking. This process has allowed us to create a more comprehensive and versatile dataset, better suited to the challenges of multimodal label relevance ranking.

Specifically, we select 3,206 clips from 219 videos in the MovieNet dataset \cite{huang2020movienet}. 
For each movie clip, we extract frames from the video and input them into the RAM model \cite{zhang2023recognize} to obtain image labels. 
Concurrently, we input the descriptions of each movie clip into the LLaMa2 model \cite{touvron2023llama} and extract correspoinding class labels.
These generated image and text labels are then filtered and modified manually, which ensures that accurate and comprehensive annotations are selected for the video clips. We also standardize each clip into 20 labels through truncation or augmentation.
As a result, we annotate 101,627 labels for 2,551 clips, with a total of 15,234 distinct label classes.
%
We refer to the new benchmark obtained from our further annotation of the MovieNet dataset as LRMovieNet (Label Relevance of MovieNet). 

\section{Experiments}
\label{sec:exp}


\begin{table*}
[t]
\begin{adjustbox}{width=0.9\textwidth, center}
\begin{tabular}{c|c|c|c|c|c|c}
\toprule
\multicolumn{2}{c|}{Method} & NDCG @ 1 & NDCG@3 & NDCG@5 & NDCG@10 & NDCG@20  \\
\midrule 
\multirow{2}{*}{OV-based} & CLIP \cite{clip}  &0.5523 &0.5209 &0.5271 &0.6009 &0.7612\\   
& MKT \cite{he2023open}  &0.3517 &0.3533 &0.3765 &0.4704 &0.6774\\   
\midrule 
\multirow{6}{*}{LTR-based} 
& PRM \cite{pei2019personalized}   &0.6320 &0.6037 &0.6083 &0.6650 &0.8022\\   
& DLCM \cite{ai2018learning}  &0.6153 &0.5807 &0.5811 &0.6310 &0.7866\\  
& ListNet \cite{cao2007learning}  &0.5947 &0.5733 &0.5787 &0.6438 & 0.7872\\  
& GSF \cite{ai2019learning}  &0.594 &0.571 &0.579 &0.643 &0.787\\  
& SetRank \cite{pang2020setrank}  &0.6337 &0.6038 &0.6125 &0.6658 &0.8030\\  
& RankFormer \cite{buyl2023rankformer}  & 0.6350 & 0.6048 & 0.6108 & 0.6655 & 0.8033\\
\midrule 
\multirow{2}{*}{Ours} 
& LR\textsuperscript{2}PPO (S1) &0.6330 &0.6018 &0.6061 &0.6667 &0.8021 \\
& LR\textsuperscript{2}PPO &\textbf{0.6820} &\textbf{0.6714} &\textbf{0.6869} &\textbf{0.7628} &\textbf{0.8475} \\
\bottomrule
\end{tabular}
\end{adjustbox}
\caption{State-of-the-art comparison for Label Relevance Ranking task on the LRMovieNet dataset. 
\textbf{Bold} indicates the best score.}
\label{tab:sota}
\end{table*}

\subsection{Experiments Setup}

\noindent\textbf{LRMovieNet Dataset}
To assess the effectiveness of our approach, we conduct experiments using the LRMovieNet dataset. After obtaining image and text labels, we split the video dataset into source and target domains based on video label types. As label relevance ranking focuses on multimodal input, we partition the domains from a label perspective. To highlight label differences, we divide the class label set by video genres. Notably, there is a significant disparity between head and long-tail genres. Thus, we use the number of clips per genre to guide the partitioning. Specifically, we rank genres by the number of clips and divide them into sets $S_P$ and $S_Q$. Set $S_P$ includes genres with more clips, while set $S_Q$ includes those with fewer clips (\ie, long-tail genres).

We designate labels in set $S_Q$ as the target domain and the difference between labels in sets $S_P$ and $S_Q$ as the source domain, achieving domain partitioning while maintaining label diversity between domains.
For source domain labels, we manually assign relevance categories (high, medium, low) based on their relevance to video clip content. For target domain labels, we randomly sample 5\%-40\% of label pairs and annotate their relative order based on their relevance to the video clip content.
To evaluate our label relevance ranking algorithm, we also annotate the test set in the target domain with high, medium, and low relevance categories for the labels.
We obtain 2551/2206/1000 video clips for the first stage/second stage/test split. The first stage data contains 10393 distinct labels, while the second stage and validation set contain 4841 different labels.


\begin{table}
[t!]
\begin{adjustbox}{width=0.8\textwidth, center}
\begin{tabular}{c|c|c|c|c|c}
\toprule
Method & NDCG @ 1 & NDCG@3 & NDCG@5 & NDCG@10 & NDCG@20  \\
\midrule 
 PRM \cite{pei2019personalized}  &0.5726
&0.5804
&0.5973
&0.6407
&0.7603\\   
DLCM \cite{ai2018learning}  &0.5983
&0.6025
&0.6125
&0.6797
&0.7744\\  
 ListNet \cite{cao2007learning}  &0.5449
&0.5575
&0.5699
&0.6324
&0.7467\\  
 GSF \cite{ai2019learning}   &0.6004
&0.6265
&0.6471
&0.7054
&0.7892\\ 
SetRank \cite{pang2020setrank}
&0.5299
&0.5380
&0.5555
&0.6083
&0.7365\\ 
RankFormer \cite{buyl2023rankformer}
&0.5684
&0.5511
&0.5643
&0.6164
&0.7458\\
\midrule 
 LR\textsuperscript{2}PPO &\textbf{0.6496}
&\textbf{0.6830}
&\textbf{0.7033}
&\textbf{0.7710}
&\textbf{0.8240} \\
\bottomrule
\end{tabular}
\end{adjustbox}
\caption{State-of-the-art comparison on traditional datasets
for label relevance ranking on the MSLR-Web10K $\rightarrow$ MQ2008 transfering task. 
}
\label{tab:web30k_mq2008}
\end{table}

\noindent\textbf{MSLR-WEB10k $\rightarrow$ MQ2008}
To demonstrate the generalizability of our method, we further conduct experiments on traditional LTR datasets.
In this transfer learning scenario, we use MSLR-WEB10k as the source domain and MQ2008 as the target domain, based on datasets introduced by Qin and Liu \cite{DBLP:journals/corr/QinL13}.


\noindent\textbf{Evaluation Metrics }
We use the NDCG (Normalized Discounted Cumulative Gain) metric as the evaluation metric for multimodal label relevance ranking.
For each video clip, we compute NDCG@$k$ for the top $k$ labels.



\subsection{State-of-the-art Comparison}

\noindent\textbf{LRMovieNet Dataset}
We compare LR\textsuperscript{2}PPO with previous state-of-the-art LTR methods, reporting NDCG metrics for the LRMovieNet dataset. 
Results of LTR-based methods are reproduced based on the paper description, since the original models can not be directly applied to this task.
As seen in Tab.~\ref{tab:sota}, LR\textsuperscript{2}PPO significantly outperforms previous methods. Meanwhile, compared to the first-stage model LR\textsuperscript{2}PPO (S1), 
the final LR\textsuperscript{2}PPO achieves consistent improvement of over 3\% at different NDCG@k. 
Open-vocabulary (OV) based methods, such as CLIP and MKT, utilize label confidence for the ranking of labels, exhibiting relatively poor performance. 
Furthermore, these models solely focus on specific objects, thus perform inadequately when dealing with semantic information. 
LTR-based methods shows superiority over OV-based ones. However, they are not originally designed for ranking for labels, especially when transferring to a new scenario.
In contrast, our LR\textsuperscript{2}PPO model allows the base model to interact with the environment and optimize over unlabeled data, resulting in superior performance compared to other baseline models. 
%
%

\begin{table*}[t!]
\centering
\begin{adjustbox}{width=0.985\textwidth}
\begin{tabular}{c|c|c|c|c|c|c}
\toprule
\multirow{2}{*}{Annotation Proportion} & \multirow{2}{*}{Reward Model Accuracy} & \multirow{2}{*}{NDCG@1} & \multirow{2}{*}{NDCG@3} & \multirow{2}{*}{NDCG@5} & \multirow{2}{*}{NDCG@10} & \multirow{2}{*}{NDCG@20} \\
& & & & & & \\
\midrule
$ 0\% $&- & 0.6330
&0.6018
&0.6061
&0.6667
&0.8021 \\
$ 5\% $&0.7697 &0.6787
&0.6581
&0.6770
&0.7514
&0.8416 \\
$ 10\% $ &0.7757 &0.6820&0.6714
&0.6869
&0.7628
&0.8475 \\
$ 20\% $&0.7837 &0.6800
&0.6784
&0.6980
&0.7667
&0.8506 \\
$ 40\% $&0.7866 &0.6830
&0.6682
 &0.6877
&0.7617
&0.8467 \\
\bottomrule
\end{tabular}
\end{adjustbox}
\caption{Stage 2 and 3 results with different annotation proportions in target domain.}
\label{tab:anno_prop_target_domain}
\end{table*}


\begin{figure}[t]
\centering
\begin{subfigure}{0.4\textwidth} 
\centering
\includegraphics[width=\textwidth, page=1]{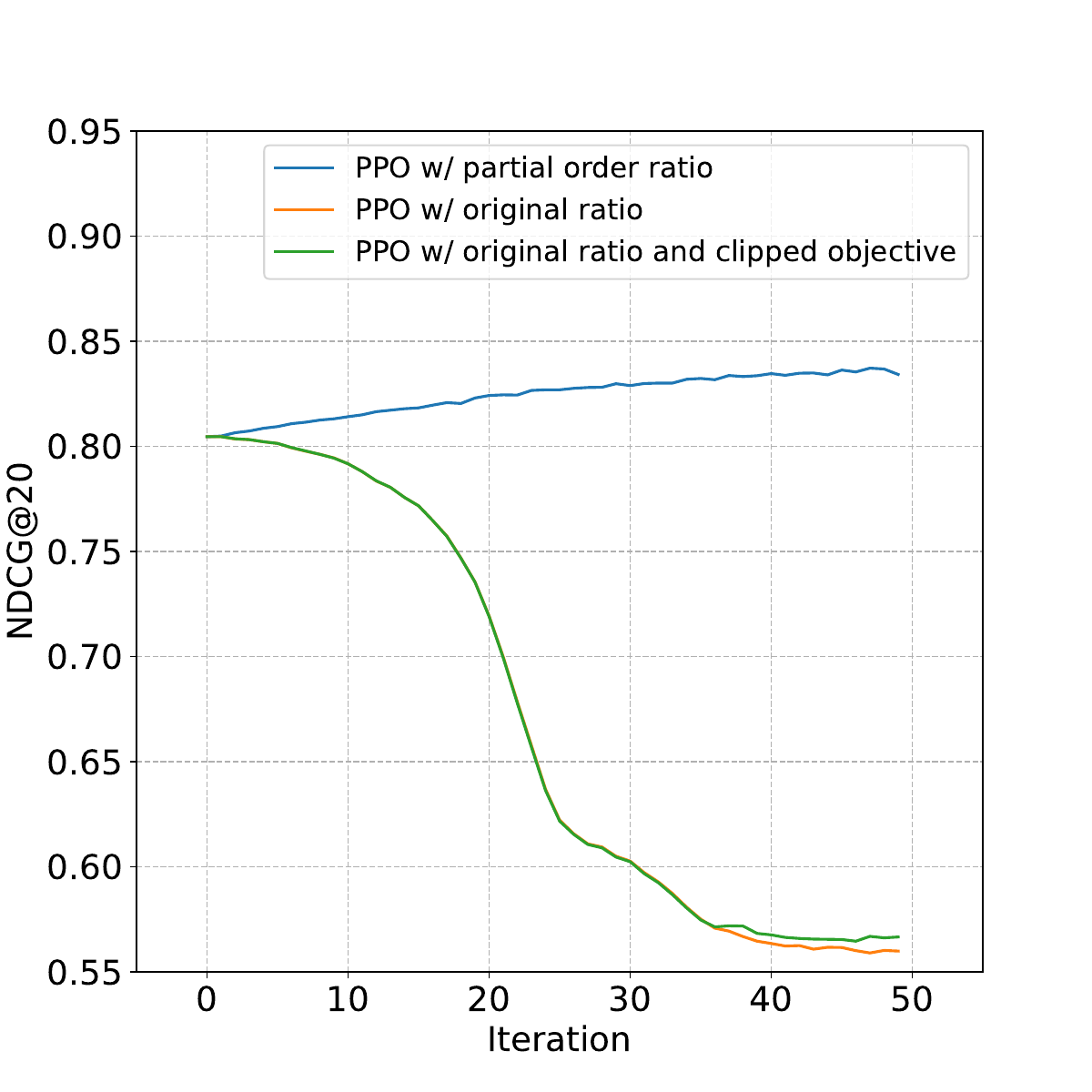}
\caption{Different ratio design}
\label{fig:ratio}
\end{subfigure}
\hspace{1cm} 
\begin{subfigure}{0.5\textwidth} 
\centering
\includegraphics[width=\textwidth, page=1]{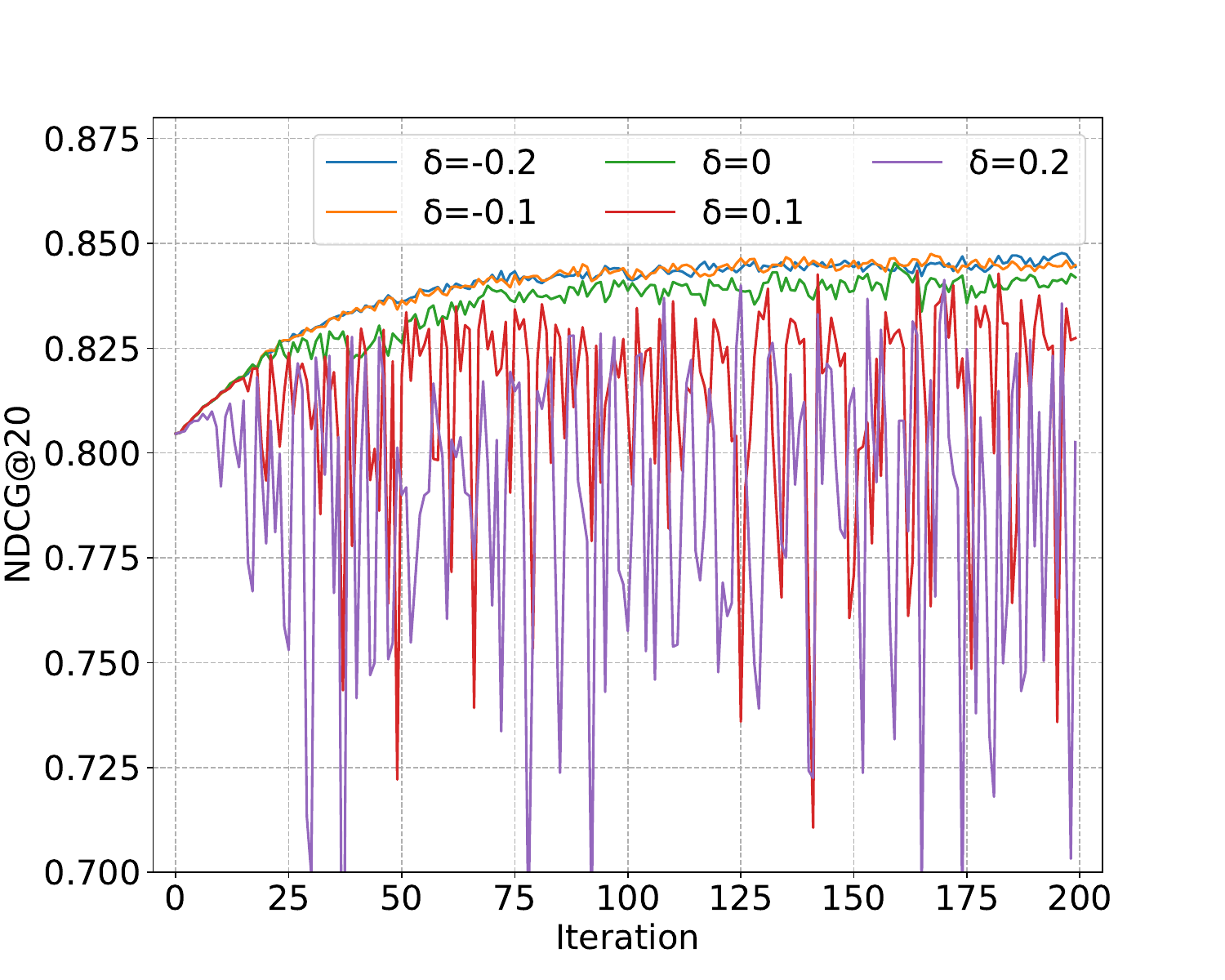}
\caption{Different threshold $\delta$}
\label{fig:threshold}
\end{subfigure}
\caption{NDCG curves during training. (a) PPO with different ratio design. Original ratio in PPO is not applicable to the definitions of state and action in the ranking task, leading to a training collapse, while our proposed partial order ratio solves this problem. (b) PPO with different thresholds $\delta$ in $r'_t(\theta)$. A small negative threshold $\delta=-0.1$ stabilizes the training, leading to superior performance.}
\label{fig:combined}
\end{figure}

\noindent\textbf{MSLR-WEB10k $\rightarrow$ MQ2008}
Comparison on LTR datasets are shown in Tab.~\ref{tab:web30k_mq2008}. 
It can concluded that our proposed LR\textsuperscript{2}PPO method surpasses previous LTR methods significantly when performing transfer learning on LTR datasets.


\subsection{Ablation Studies}
\label{sec:ablation}

\noindent\textbf{Annotation Proportion in Target Domain }
To explore the influence of the annotation proportion of ordered pairs in the target domain, we adjust the annotation proportion during the training of the reward model in stage 2. Subsequently, we adopt the adjusted reward model when training the entire LR\textsuperscript{2}PPO framework in stage 3.
The accuracy of the reward model in the second stage and the NDCG metric in the third stage are reported in Tab.~\ref{tab:anno_prop_target_domain}. 
The proportion of 10\% achieves relatively high reward model accuracy and ultimate ranking relevance, while maintaining limited annotation.

\noindent\textbf{Partial Order Ratio }
As shown in Fig.~\ref{fig:ratio}, partial order ratio shows stability in training and achieves better NDCG, 
in contrast to the original ratio in PPO, which experiences a training collapse. This result indicates that the original ratio in PPO may not be directly applicable in the setting of label relevance ranking, thereby demonstrating the effectiveness of our proposed partial order ratio. 

\noindent\textbf{Hyper-parameters Sensitivity }
Here, we investigate the impact of the threshold $\delta$ in Eq.~(\ref{eq:r_t_prime}). 
As shown in Fig.~\ref{fig:threshold}, negative thresholds $-0.1$ achieves better NDCG scores, while improving the robustness of training procedure in stage 3. Refer to supplementary for more details.


\subsection{Qualitative Assessment}

\begin{figure*}[t]
    \centering
\subcaptionbox*{}
    {
    \includegraphics[width=0.47\textwidth, page=1]{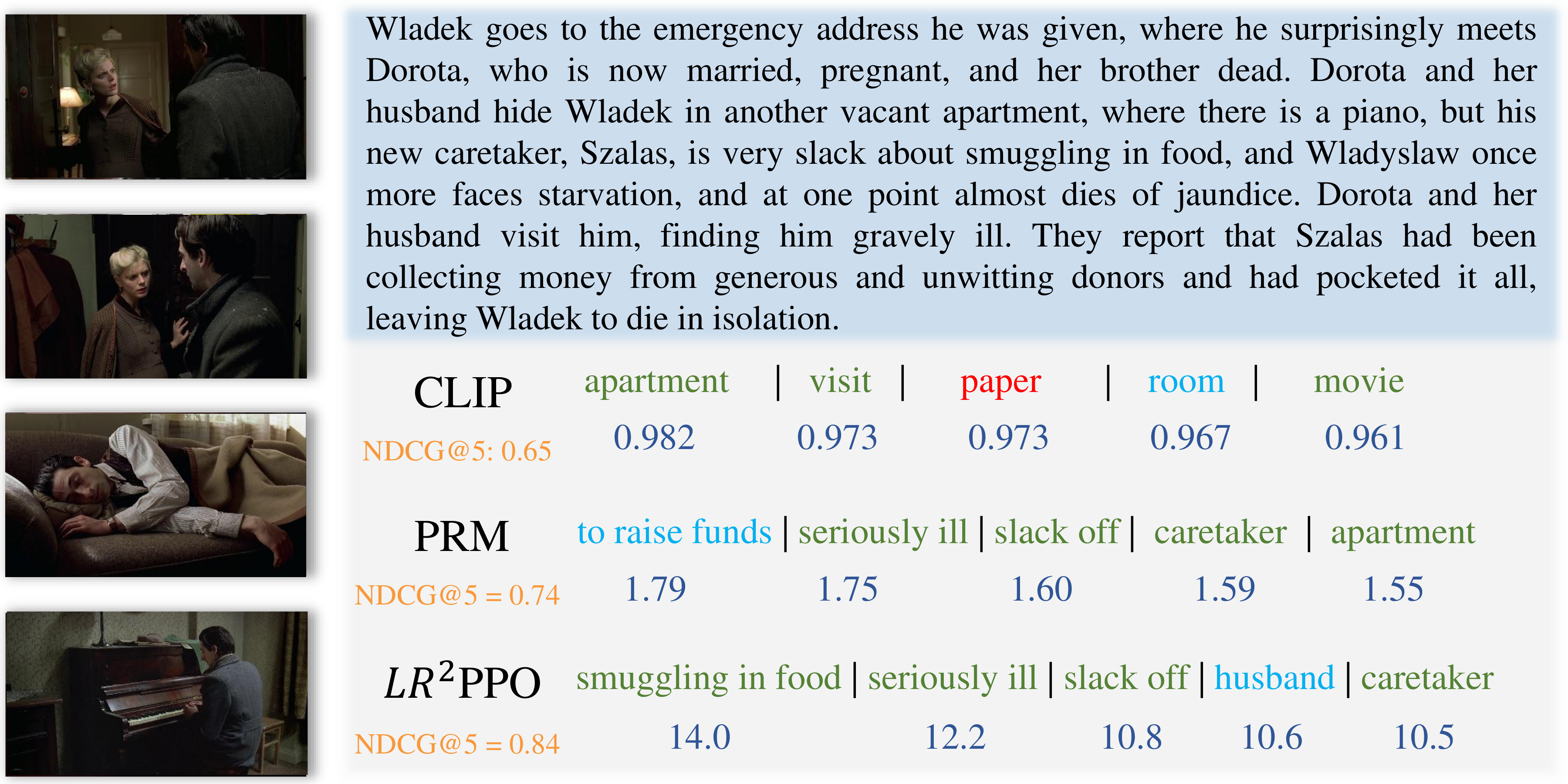}
    }
    \subcaptionbox*{}
    {
    \includegraphics[width=0.47\textwidth, page=2]{fig/qualitative.pdf}
    }
    \caption{Comparison between LR\textsuperscript{2}PPO and other state-of-the-art ranking methods.
    The \textcolor[rgb]{1,0,0}{red},  \textcolor[rgb]{0,0.690,0.941}{blue} and \textcolor[rgb]{0.329,0.510,0.208}{green} labels listed after the method represent low, medium and high in ground truth, respectively.
    The \textcolor[rgb]{0.1843, 0.3333, 0.5922}{value} below each label represents the corresponding relevance score. Best viewed in color and zoomed in.
    \label{fig:qualitative}
    }
\end{figure*}

To clearly reveal the effectiveness of our method, we visualize the label relevance ranking prediction results of the LR\textsuperscript{2}PPO algorithm and other state-of-the-art OV-based or LTR-based methods on some samples in the LRMovieNet test set.
Fig.~\ref{fig:qualitative} shows the comparison between the LR\textsuperscript{2}PPO algorithm and CLIP and PRM.
For a set of video frame sequences and a plot text description, as well as a set of labels, we compare the ranking results of different methods based on label relevance for the given label set, and list the top 5 high relevance labels predicted by each method. 
Compared with CLIP and PRM, our method ranks more high relevance labels at the top and low relevance labels at the bottom.
The results show that our method can better rank the labels based on the relevance between the label and the multimodal input, to more accurately obtain high-value labels.

\section{Conclusion}
\label{sec:conclusion}

In this study, we prove the pivotal role of label relevance in label tasks, and propose a novel approach, named LR\textsuperscript{2}PPO, to effectively mine the partial order relations and apply label relevance ranking, especifically for a new scenario. 
To evaluate the performance of the method, a new benchmark dataset, named LRMovieNet, is proposed.
Experimental results on this dataset and other LTR datasets validate the effectiveness of our proposed method.

%
%
\bibliographystyle{splncs04}
\bibliography{main}

\setcounter{page}{1}

\appendix 

\renewcommand\thesection{\Alph{section}}
\renewcommand\thefigure{\Alph{figure}}
\renewcommand\thetable{\Alph{table}}

\numberwithin{equation}{section} 
\numberwithin{algorithm}{section} 
\counterwithin{figure}{section} 
\counterwithin{table}{section} 




\textbf{\renewcommand{\thealgorithm}{1}}


\section{PPO Algorithm}

In this section, we provide a detailed explanation of the original Proximal Policy Optimization (PPO) algorithm, as proposed by Schulman et al. \cite{schulman2017proximal}. 
The PPO algorithm is defined by three key loss functions: the policy function loss, the value function loss, and the entropy bonus.
In Sec.~\hyperlink{fake}{3.2} of the main paper, 
we present the concept of the target value function estimate $V_t^\text{{target}}$ and the advantage estimate $\hat{A}_t$ at time step $t$.
Building upon this, we proceed to introduce the details of the PPO algorithm.

Formally, $r_t(\theta)$ is defined as the ratio of the action probability under the policy $\pi_\theta$ to the action probability under the old policy $\pi_{\theta_{\text{old}}}$:
\begin{equation}
\label{eq:ppo_ratio}
r_t(\theta) = \frac{\pi_\theta(a_t|s_t)}{\pi_{\theta_{\text{old}}}(a_t|s_t)}.
\end{equation}
Subsequently, the policy function loss, denoted as $L^\text{PF}(\theta)$, is computed based on the policy parameters $\theta$:
\begin{equation}
\label{eq:ppo_policy_loss}
L^{PF}(\theta) = - \mathbb{E}_t \left( r_t(\theta) \hat{A}_t \right).
\end{equation}
To prevent the policy from changing too drastically in a single update, the PPO algorithm typically employs a clipped formation of policy loss:
\begin{align}
\label{eq:ppo_clipped_policy_loss}
L^{CPF}(\theta) = - \mathbb{E}_t \Bigl[ \min \Bigl( r_t (\theta) \hat{A}_t, \text{clip}\bigl( r_t(\theta),  1-\epsilon, 1+\epsilon \bigr) \hat{A}_t \Bigr) \Bigr].
\end{align}
The value function loss, denoted as $L^{VF}(\omega)$, has also been defined 
in Sec.~\hyperlink{fake}{3.2} of the main paper.

The entropy bonus, denoted as $S(\pi_\theta)$, is defined as:
\begin{equation}
\label{eq:entropy_bonus}
S(\pi_\theta) = \mathbb{E}_t \left[ \sum_{a_t} - \pi_\theta(a_t|s_t) \log \pi_\theta(a_t|s_t) \right] .
\end{equation}
This is the expected value at time step $t$ of the sum over all possible actions of the product of the action probability under the policy $\pi_\theta$ and the logarithm of the action probability under the policy $\pi_\theta$. 
This term encourages exploration by maximizing the entropy of the policy.
Ultimately, the total loss of PPO can be defined as:
\begin{equation} 
L'(\theta) = L^{CPF}(\theta) + c_1' L^{VF}(\omega) - c_2' S(\pi_\theta),  
\end{equation}
where  $c_1'$ and $c_2'$ represent adjustable hyperparameters, which can be tuned to optimize the performance of the model.

In many cases, instead of using the clipped policy loss form in Eq.~(\ref{eq:ppo_clipped_policy_loss}), the PPO algorithm incorporates a KL penalty term in the overall loss to prevent overly large policy updates that could lead to instability or performance drops in the learning process. 
The KL penalty, denoted as $KL_{penalty}$, is formulated as:
\begin{equation}
\label{eq:kl_penalty}
    KL_{penalty} (\pi_{\theta_{old}}, \pi_{\theta}) = E_t[KL(\pi_{\theta_{old}}(a_t|s_t), \pi_\theta(a_t|s_t))],
\end{equation}
where $KL(\cdot, \cdot)$ represents the Kullback-Leibler (KL) divergence.
Thereby, the overall loss function, denoted as $L''(\theta)$, is a combination of the four aforementioned loss functions. It is computed as:
\begin{align} 
L''(\theta) = L^{PF}(\theta) + c_1'' L^{VF}(\omega) - c_2'' S(\pi_\theta) + c_3'' KL_{penalty} (\pi_{\theta_{old}}, \pi_{\theta})  .
\end{align}
Here, the hyperparameters $c_1''$,$c_2''$ and $c_3''$ are used to balance the contributions of the four components to the overall loss, and are typically determined through empirical tuning.

\section{LR\textsuperscript{2}PPO Algorithm}

The joint training of Label Relevance Ranking with Proximal Policy Optimization (LR\textsuperscript{2}PPO) framework is illustrated in Algorithm~\ref{alg:lr2ppo}.
Note that the definitions of entropy bonus $S(\pi_\theta) $ and KL penalty $KL_{penalty} (\pi_{\theta_{old}}, \pi_{\theta}) $ are the same as the original PPO. (See Eq.~(\ref{eq:entropy_bonus}) and Eq.~(\ref{eq:kl_penalty}).)

\begin{algorithm}
\caption{Label Relevance Ranking with Proximal Policy Optimization (LR\textsuperscript{2}PPO), Full Procedure}
\label{alg:lr2ppo}
\begin{algorithmic}[1]
\REQUIRE $T=$ Maximal state transition timestep length, 
$N_{\text{Trajs}}=$ Number of state transition trajectories collected as training data in a training iteration, 
$K=$ Number of learning epochs in a training iteration, 
$N_{\text{Iters}}=$ Number of training iterations, weighting factors $c_1$, $c_2$ and $c_3$, 
$M=$ Minibatch size, $\gamma=$ Discount factor through timesteps, 
$m=$ Margin in partial order function,
$\delta=$ Threshold when calculating partial order ratio
\ENSURE  $\pi_\theta$, $V_{\omega}$
\STATE $\pi_\theta \leftarrow$ newActorNet (), $V_\omega \leftarrow$ newCriticNet (), optimizer $\leftarrow$ newOptimizer $\left(\pi_\theta, V_\omega\right)$
\STATE env $\leftarrow$ Muitimodal input and corresponding labels \hfill $\triangleright$ Environment of label relevance ranking task
\STATE num\_batches = $\lfloor \frac{N_{\text{Trajs}} \cdot T}{M} \rfloor$ \hfill $\triangleright$ Number of batches in a training iteration
\FOR{$n_{iter}=1,2, \ldots, {N_{\text{Iters}}}$}
    \STATE train\_data $\leftarrow[]$
    \newline
    // Produce training data.
    \FOR{$n_k=1,2,\ldots,{N_{\text{Trajs}}}$}
        \STATE $s_{t=0} \leftarrow$ env.randomlySampleLabelPair () \hfill $\triangleright$ Randomly sample pair of labels as initial state
        \newline
        // Let actor interact with environment and collect training data:
        \FOR{$t=1,2, \ldots, {T}$}
            \STATE $a_t \leftarrow \pi_\theta$. generate\_action $\left(s_t\right)$
            \STATE $\pi_{\theta_{\text {old }}}\left(a_t \mid s_t\right) \leftarrow \pi_{\theta}.\text { distribution.get\_probability }\left(a_t\right)$ \hfill $\triangleright$ Old action logits
            \STATE $s_{t+1}, r_t \leftarrow$ env.step $\left(a_t\right)$ \hfill $\triangleright$ Transition state and get reward
            \STATE train\_data $\leftarrow$ train\_data $+\operatorname{tuple}\left(s_t, a_t, r_t, \pi_{\theta_{\text {old }}}\left(a_t \mid s_t\right)\right)$
        \ENDFOR
        \newline
        // Calculate and collect target value and advantage at timestep $t$
        \FOR{$t=1,2, \ldots, {T}$}
            \STATE $V_{\omega_{old}} (s_T) \leftarrow V_{\omega}$.get\_value($s_T$)
            \STATE $V_{\omega_{old}} (s_t) \leftarrow V_{\omega}$.get\_value($s_t$)
            \STATE $V_t^{\text {target }}=r_t+\gamma r_{t+1}+\gamma^2 r_{t+2}+\ldots+\gamma^{T-t-1} r_{T-1}+\gamma^{T-t} V_{\omega_{old}}\left(s_T\right)$ \hfill $\triangleright$ Target Value
            \STATE $\hat{A}_t=V_t^{\text {target }}-V_{\omega_{old}}\left(s_t\right)$ \hfill $\triangleright$ Estimated advantage
            \STATE train\_data $[(n_k-1)\cdot T + t] \leftarrow$ train\_data $[(n_k-1)\cdot T + t]+\operatorname{tuple}\left(A_t, V_t^{\text {target }}\right)$
        \ENDFOR
    \ENDFOR
    \STATE optimizer.resetGradients $\left(\pi_\theta, V_\omega\right)$
    \newline
    // Update trainable parameters $\theta$ and $\omega$ for $E$ epochs:
    \FOR{epoch $=1,2, \ldots, K$}
        \STATE train\_data $\leftarrow$ randomizeOrder(train\_data)
        \FOR{batch\_index $=1,2, \ldots$, num\_batches}
            \STATE $\mathrm{E_{batch}} \leftarrow \text { getNextBatch(train\_data, batch\_index, M) }$ \hfill $\triangleright$ Get minibatch for training
            \FOR{example $\mathrm{e} \in \mathrm{E_{batch}}$}
                \STATE $s_t, a_t, r_t, \pi_{\theta_{\text {old }}}\left(a_t \mid s_t\right), \hat{A}_t, V_t^{\text {target }} \leftarrow \operatorname{unpack}(e)$
                \STATE $\_\leftarrow \pi_\theta$.generate\_action $\left(s_t\right)$ \hfill $\triangleright$ Parameterization of policy
                \STATE ${\pi}_\theta\left(a_t \mid s_t\right) \leftarrow \pi_\theta$. distribution.get\_probability $\left(s_t\right)$ \hfill $\triangleright$ Action logits
                \STATE $p_t^1,p_t^2 \leftarrow \operatorname{unpack} ( {\pi}_\theta\left(a_t \mid s_t\right) )$ \hfill $\triangleright$ Action\_logit for each label in the label pair

                \algstore{myalg}
                \end{algorithmic}
                \end{algorithm}
                
                \begin{algorithm}                     
                \begin{algorithmic} [1]                   
                \algrestore{myalg}
                
                \STATE $ r_t^{\prime}(\theta) = \begin{cases} - \operatorname{max}(0,m-(p_t^1-p_t^2)), \hat{A}_t \geq \delta \\     - \operatorname{max}(0,m-(p_t^2-p_t^1))  ,\hat{A}_t < \delta\end{cases} $ \hfill $\triangleright$ Partial order ratio
            \ENDFOR
            \STATE $L_{LR\textsuperscript{2}PPO}^{PF}(\theta) = - \frac{1}{|M|} \sum_{t \in\{1,2, \ldots,|M|\}} r^{\prime}_{t}(\theta) \operatorname{abs} (\hat{A}_t) $ \hfill $\triangleright$ Policy loss
            \STATE $L_ {LR\textsuperscript{2}PPO}^{VF} (\omega) = \frac{1}{|M|} \sum_{t \in\{1,2, \ldots,|M|\}} \left( V_\omega(s_t) - V^{target}_t \right)^2 $ \hfill $\triangleright$ Value loss
            \STATE $S(\pi_\theta) = -\frac{1}{|M|} \sum_{t \in{1,2, \ldots,|M|}} \sum_{a_t} \pi_\theta(a_t|s_t) \log \pi_\theta(a_t|s_t)  $ \hfill $\triangleright$ Entropy bonus 
            \STATE $KL_{penalty} (\pi_{\theta_{old}}, \pi_{\theta}) = \frac{1}{|M|} \sum_{t \in{1,2, \ldots,|M|}}KL(\pi_{\theta_{old}}(a_t|s_t), \pi_\theta(a_t|s_t))$ \hfill $\triangleright$ KL penalty
            \STATE $L_{LR\textsuperscript{2}PPO}(\theta, \omega) = L_{LR\textsuperscript{2}PPO}^{PF}(\theta) + c_1 L_{LR\textsuperscript{2}PPO}^{VF}(\omega) - c_2 S(\pi_\theta) + c_3 KL_{penalty} (\pi_{\theta_{old}}, \pi_{\theta}) $ \hfill $\triangleright$ Total loss
            \STATE optimizer.backpropagate $(\pi_\theta, V_\omega, L_{LR\textsuperscript{2}PPO}(\theta, \omega))$
            \STATE optimizer.updateTrainableParameters $\left(\pi_\theta, V_\omega\right)$
        \ENDFOR
    \ENDFOR
\ENDFOR
\RETURN $\pi_\theta$, $V_{\omega}$
\end{algorithmic} 
\end{algorithm}


\section{Experiments}

\subsection{More Details about LRMovieNet Dataset}


\noindent\textbf{Prompts to Extract Text Labels}
When generating textual labels from scene description in each movie clip of MovieNet~\cite{huang2020movienet} dataset, we feed the following prompts into LLaMa2~\cite{touvron2023llama} to obtain 
event labels and entity labels, respectively.
(1) Event labels:
``Below is an instruction that describes a task. Write a response that appropriately completes the request. \#\#\# Instruction: Use descriptive language tags consisting of less than three words each to capture events without entities in the following sentence: \{sentence\} Please seperate the labels by numbers and DO NOT return sentence. \#\#\# Response:''
(2) Entity labels:
``Below is an instruction that describes a task. Write a response that appropriately completes the request. \#\#\# Instruction: Use descriptive language tags consisting of less than three words each to capture entities in the following sentence: \{sentence\} Please seperate the labels by numbers and DO NOT return sentence. \#\#\# Response: ''.
The produced text labels and image labels (by RAM model~\cite{zhang2023recognize}), 
are subsequently screened and manually adjusted.

\begin{figure*}[t]
\centering
\subcaptionbox{Clips count in different genres\label{fig:1a}}
    {\includegraphics[width=0.3\linewidth]{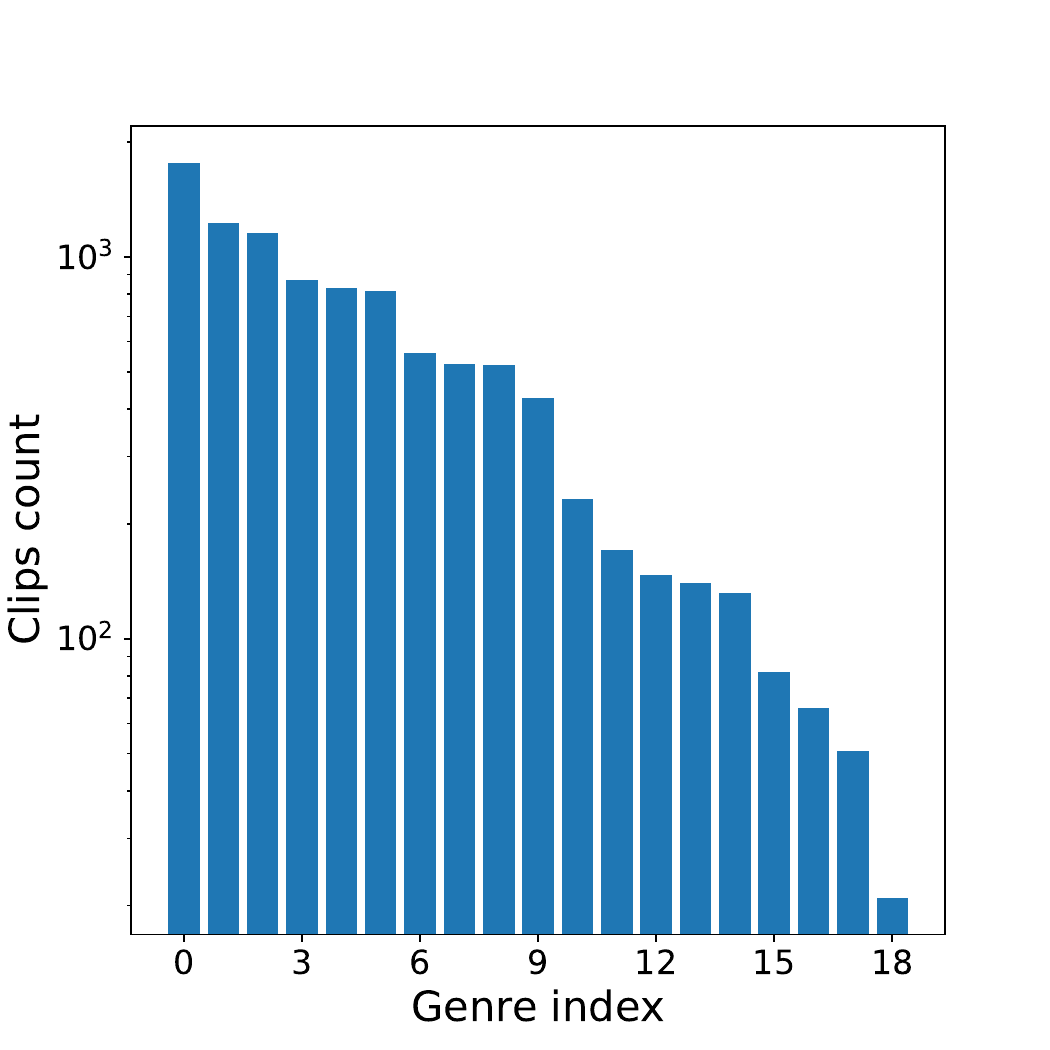}}
\subcaptionbox{Label classes count in different genres\label{fig:1b}}
    {\includegraphics[width=0.3\linewidth]{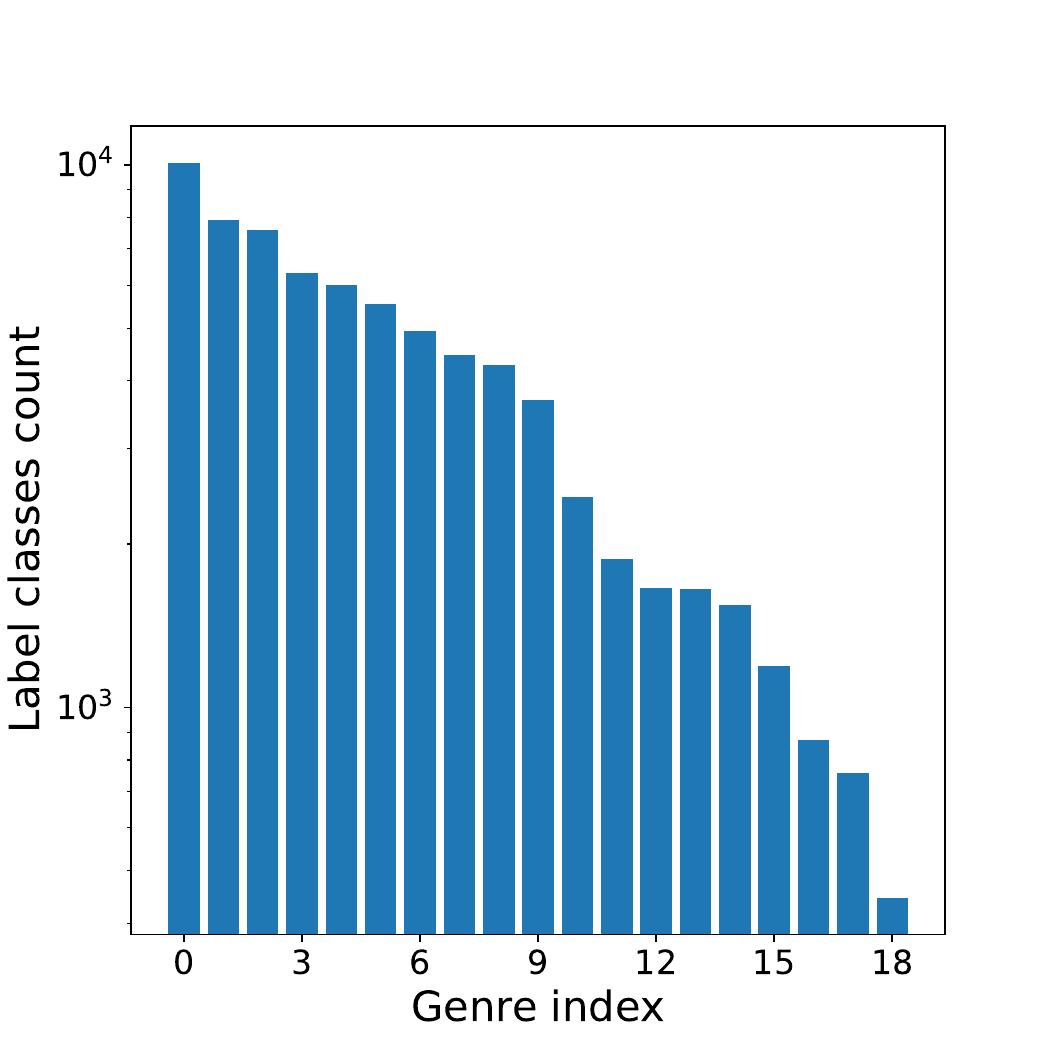}}
\subcaptionbox{Labels count about label frequency\label{fig:1c}}
    {\includegraphics[width=0.3\linewidth]{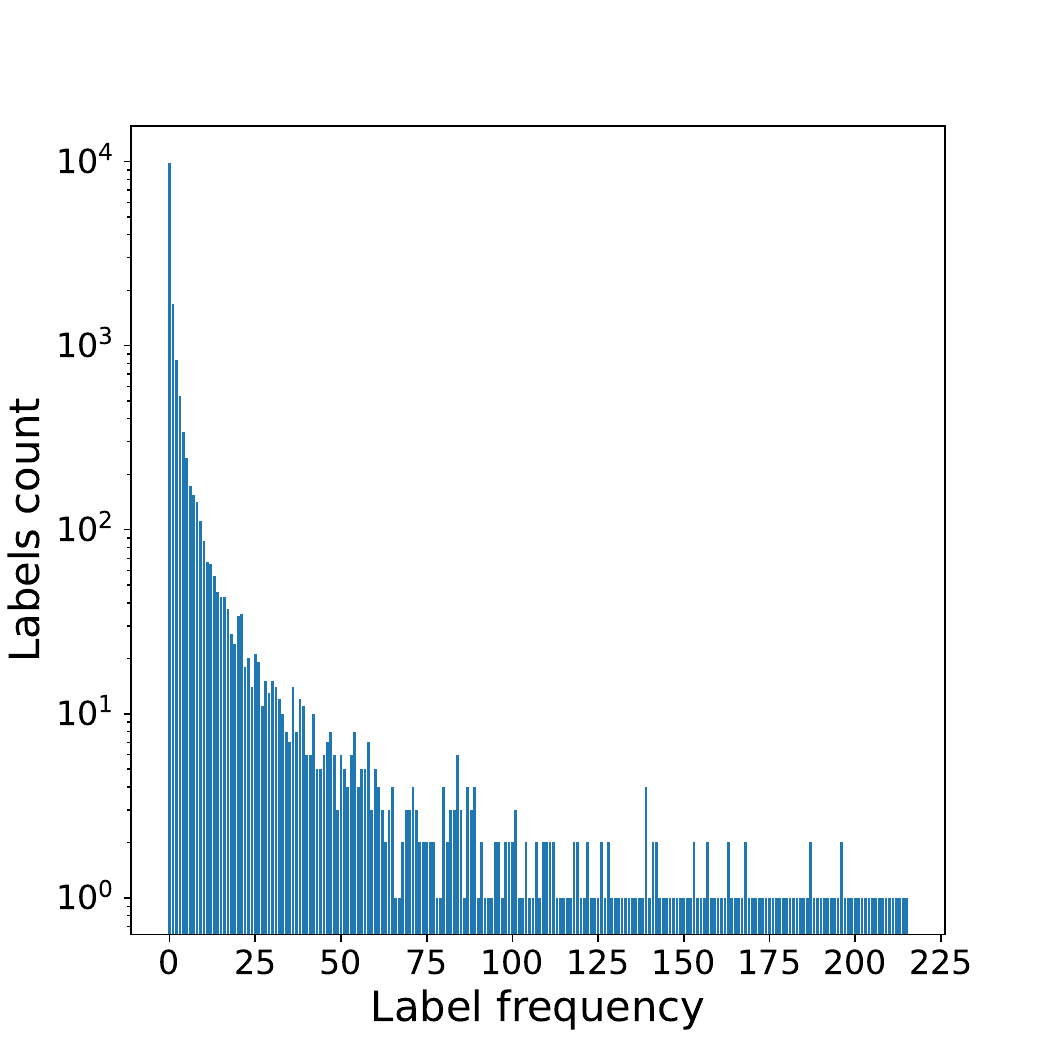}}\\
\caption{Data statistics of LRMovieNet.}
\label{fig:dataset_statistics}
\end{figure*}

\noindent\textbf{Statistics of LRMovieNet Dataset}
To better understand the overall distribution of LRMovieNet, we provide histograms of its statistics in several dimensions, 
which are displayed in Fig.~\ref{fig:dataset_statistics}.
Specifically, Fig.~\ref{fig:dataset_statistics} (a) shows the total number of video clips in all videos of different genres, Fig.~\ref{fig:dataset_statistics} (b) displays the number of label classes in different genres, 
and Fig.~\ref{fig:dataset_statistics} (c) provides the number of labels with different frequencies in the entire LRMovieNet dataset.
The details of Fig.~\ref{fig:dataset_statistics} (a) and Fig.~\ref{fig:dataset_statistics} (b) are listed in 
Tab.~\ref{subtab:genre_index_clip_count} and Tab.~\ref{subtab:genre_index_label_classes_count}, respectively.

\begin{table*}[hbt!]
\centering
\small
\begin{subtable}{.45\textwidth}
\centering
\fontsize{7pt}{8pt}\selectfont 
\begin{tabular}{c|c|c}
\toprule
\multirow{2}{*}{Genre Index} & \multirow{2}{*}{Genre Name} & \multirow{2}{*}{Clips Count} \\
&&\\
\midrule
$ 0 $ & Drama & 1765 \\
$ 1 $ & Action & 1224\\
$ 2 $ & Thriller & 1154\\
$ 3 $ & Sci-Fi & 869 \\
$ 4 $ & Crime & 829\\
$ 5 $ & Adventure & 814 \\
$ 6 $ & Comedy & 562\\
$ 7 $ & Mystery & 525 \\
$ 8 $ & Fantasy & 520 \\
$ 9 $ &Romance & 427\\
$ 10 $ & Biography & 232\\
$ 11 $ & War & 171\\
$ 12 $ &Horror & 147\\
$ 13 $ & Family & 140 \\
$ 14 $ &History & 132 \\
$ 15 $ &Music & 82 \\
$ 16 $ &Western & 66 \\
$ 17 $ & Sport & 51\\
$ 18 $ & Musical & 21 \\
\bottomrule
\end{tabular}
\caption{Details about clips count in different genres}
\label{subtab:genre_index_clip_count}
\end{subtable}
\hfill
\begin{subtable}{.45\textwidth}
\centering
\fontsize{7pt}{8pt}\selectfont 
\begin{tabular}{c|c|c}
\toprule
\multirow{2}{*}{Genre Index} & \multirow{2}{*}{Genre Name} & \multirow{2}{*}{Classes Count} \\
&&\\
\midrule
$ 0 $ &  Drama & 10088 \\
$ 1 $ &  Action & 7923 \\
$ 2 $ &  Thriller & 7573 \\
$ 3 $ &  Sci-Fi & 6315 \\
$ 4 $ &  Adventure & 5999 \\
$ 5 $ &  Crime & 5542 \\
$ 6 $ &  Comedy & 4933 \\
$ 7 $ &  Fantasy & 4468 \\
$ 8 $ &  Mystery & 4277 \\
$ 9 $ &  Romance & 3688 \\
$ 10 $ &  Biography & 2443 \\
$ 11 $ &  War & 1879 \\
$ 12 $ &  Horror & 1658 \\
$ 13 $ &  History & 1652 \\
$ 14 $ &  Family & 1543 \\
$ 15 $ &  Music & 1191 \\
$ 16 $ &  Western & 872 \\
$ 17 $ &  Sport & 757 \\
$ 18 $ &  Musical & 446 \\
\bottomrule
\end{tabular}
\caption{Details about label classes count in different genres}
\label{subtab:genre_index_label_classes_count}
\end{subtable}
\caption{Details about number of video clips and label classes in all videos of different genres in LRMovieNet.}
\label{tab:genre_index_details}
\end{table*}

\begin{figure*}[t]
\centering
\begin{tabular}{cc}
\multicolumn{2}{c}{
(a) Source Domain
} \\
\includegraphics[width=0.47\textwidth, page=1]{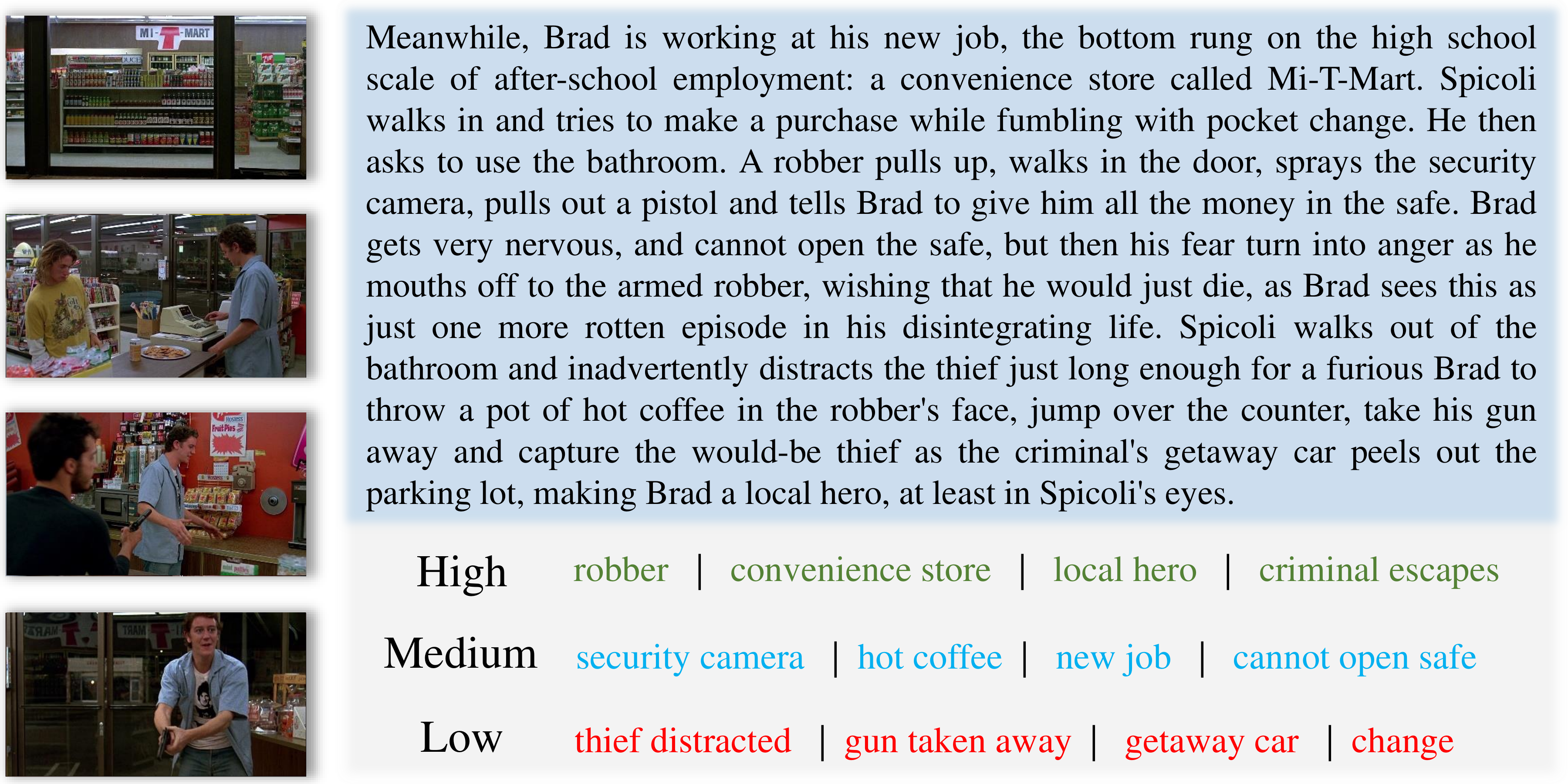} &
\includegraphics[width=0.47\textwidth, page=2]{fig/qualitative_supp.pdf} \\
\multicolumn{2}{c}{
(b) Target Domain
} \\
\includegraphics[width=0.47\textwidth, page=3]{fig/qualitative_supp.pdf} &
\includegraphics[width=0.47\textwidth, page=4]{fig/qualitative_supp.pdf}
\end{tabular}
\caption{
Annotated training samples in source and target domains.
The \textcolor[rgb]{1,0,0}{red},  \textcolor[rgb]{0,0.690,0.941}{blue} and \textcolor[rgb]{0.329,0.510,0.208}{green} labels listed in the upper subfigure represent low, medium and high in ground truth in the source domain, respectively.
For each label pair in the lower subfigure, the \textcolor[rgb]{0.7176470588235294, 0.6, 1.0}{left} label are more relevant than the \textcolor[rgb]{0.8509803921568627, 0.5137254901960784, 0.521568627450980}{right} in accordance with the video episode context (\ie, descriptions and frames).
Best viewed in color and zoomed in. 
\label{fig:qualitative_supp}
}
\end{figure*}

\noindent\textbf{Samples of LRMovieNet Dataset}
To illustrate the variation of types (\eg, objects, attributes, scenes, character identities) and semantic levels (\eg, general, specific, abstract) in LRMovieNet, we provide some training samples in Fig.~\ref{fig:qualitative_supp}. These samples also serve to clarify the process of label relevance categories annotation and the annotated partial order label pairs. The annotated label relevance categories in the source domain are used to train the relevance ranking base model in the first stage. Subsequently, the partial order label pairs in the target domain (along with label pairs augmented from the source domain) are employed to train the reward model in the second stage. This reward model then guides the joint training of LR\textsuperscript{2}PPO in the third stage.

\subsection{Metrics}

In this paper, we evaluate the performance of our label relevance ranking algorithm using the Normalized Discounted Cumulative Gain (NDCG) metric, which is widely adopted in information retrieval and recommender systems. 

To better comprehend the NDCG metric, let's first introduce the concept of relevance scores. These scores, denoted as $rel_i$, are assigned to the item (label) at position $i$. Based on their relevance levels, the manually annotated labels in the test set are assigned these scores: labels with high, medium, and low relevance are given scores of 2, 1, and 0, respectively.
Next, we define the Discounted Cumulative Gain (DCG) at position $k$ as follows:
\begin{equation}
\text{DCG}@k = \sum_{i=1}^{k} \frac{2^{rel_i} - 1}{\log_2{(i+1)}} ,
\end{equation}
where $rel_i$ is the relevance score of the item at position $i$. The DCG metric measures the gain of a ranking algorithm by considering both the relevance of the items and their positions in the ranking.

To obtain the Ideal Discounted Cumulative Gain (IDCG) at position $k$, we first rank the items by their relevance scores in descending order and then calculate the DCG for this ideal ranking:
\begin{equation}
\text{IDCG}@k = \sum_{i=1}^{k} \frac{2^{rel^*_i} - 1}{\log_2{(i+1)}} ,
\end{equation}
where $rel_i^*$ is the relevance score of the item (label) at position $i$ in the ideal ranking. The IDCG represents the maximum possible DCG for a given query.

Finally, we compute the Normalized Discounted Cumulative Gain (NDCG) at position $k$ by dividing the DCG by the IDCG to ensure that it lies between 0 and 1:
\begin{equation}
\text{NDCG}@k = \frac{\text{DCG}@k}{\text{IDCG}@k} .
\end{equation}
A value of 1 indicates a perfect ranking, while a value of 0 indicates the worst possible ranking. The normalization also allows for the comparison of NDCG values across different video clips, as it accounts for the varying number of labels.


\subsection{Implementation Details}

We leverage published Vision Transformer \cite{vit} and Roberta \cite{liu2019roberta} weights to initialize the parameters of vision encoder and language encoder, respectively. 
The parameters of the two encoders are fixed during three training stages.
The coefficients $c_1$ and $c_2$ in Eq.~(\hyperlink{fake}{11}) 
are set to $1$, $1\times10^{-3}$, respectively.
The coefficient $c_3$ of KL penalty in Eq.~(\hyperlink{fake}{11}) 
is set to $1\times10^{-3}$.
In our implementation, the KL penalty is subtracted from the reward, instead of being directly included in the joint loss.
In this way, the policy is still updated based on the expected return, but the return is adjusted based on the policy change.
This allows the algorithm to explore more freely while still being penalized for large policy changes, leading to more effective exploration and potentially higher overall returns.
$\gamma$ in the definition of $V_t^{target}$ is set to $0$.
$\beta$ in Eq.~(\hyperlink{fake}{3})  is set to $0.3$.
Margin $m_R$ in Eq.~(\hyperlink{fake}{4}) and margin $m$ in Eq.~(\hyperlink{fake}{7})  are both set to $1$,
and $\delta$ in Eq.~(\hyperlink{fake}{8}) is set to $-0.1$.
AdamW optimizer is adopted for all optimized networks in three training stages, with learning rate $2\times10^{-5}$ in the first and second training stages and $1\times10^{-3}$ in the third training stage.
We train the first two stages for 15 epochs, respectively.
In Algorithm~\ref{alg:lr2ppo}, $T$ is set to $1$, 
$N_{\text{Trajs}}$ is set to $200$, $K$ is set to $1$, 
$N_{\text{Iters}}$ is set to $412$, $M$ is set to $24$.
During the training of stage 2, only 10\% of ordered annotation pairs is used. 
In stage 3, 40\% of randomly sampled pairs without annotation is utilized to train the LR\textsuperscript{2}PPO framework, with the guidance of the reward model trained in stage 2.
We use four V100 GPUs, each with 32GB of memory, to train the models.
In comparison experiments, we utilize ``There is a \{label\} in the scene'' as textual prompt for both CLIP~\cite{clip} and MKT~\cite{he2023open}. 


\subsection{More Ablation Studies about LR\textsuperscript{2}PPO}
\noindent{\textbf{Classification and regression.}}
The relevance between each label and the multimodal input are annotated into high, medium and low 
(with scores 2, 1, 0, respectively).
We adopt regression instead of classification for better performance and convenience in the latter stages. The results of classification in stage 1 (S1-CLS) are listed in Tab.~\ref{tab:movienet_supp}.
We contribute the performance gap between classification (S1-CLS) and regression (S1) to the need to \textit{fully} rank in the task. 
The predicted logits of classification need to be weighted to form the final relevance score, which hinders its performance.
Regression scores are more suitable for ranking compared to classification logits.

\noindent{\textbf{Results w/o stage 1.}}
The results of omitting stage 1 (w/o S1) are listed in Tab.~\ref{tab:movienet_supp}.
LR\textsuperscript{2}PPO achieves better performance since it starts learning from the source domain pretrained model, whose parameters more coincide with the ranking task, while LR\textsuperscript{2}PPO (w/o S1) starts from the officially pretrained ViT~\cite{vit} and Roberta~\cite{liu2019roberta}.

\begin{table}
[hbt!]
\begin{adjustbox}{width=0.8\columnwidth, center}
\begin{tabular}{c|c|c|c|c|c}
\toprule
Method & NDCG @ 1 & NDCG@3 & NDCG@5 & NDCG@10 & NDCG@20  \\
\midrule 
LR\textsuperscript{2}PPO (S1-CLS) &0.6077 &0.5988 &0.5998 &0.6601 &0.7981\\
LR\textsuperscript{2}PPO (S1) &0.6330 &0.6018 &0.6061 &0.6667 &0.8021 \\
LR\textsuperscript{2}PPO (w/o S1) &0.6750 &0.6583 &0.6781 &0.7529 &0.8432 \\
 LR\textsuperscript{2}PPO &\textbf{0.6820} &\textbf{0.6714} &\textbf{0.6869} &\textbf{0.7628} &\textbf{0.8475} \\
\bottomrule
\end{tabular}
\end{adjustbox}
\caption{More ablation studies on LRMovieNet.
}
\label{tab:movienet_supp}
\end{table}


\subsection{More Hyper-parameters Sensitivity Analysis}

\begin{figure*}[!htb]
\centering
\subcaptionbox*{}
    {\includegraphics[width=0.3\linewidth]{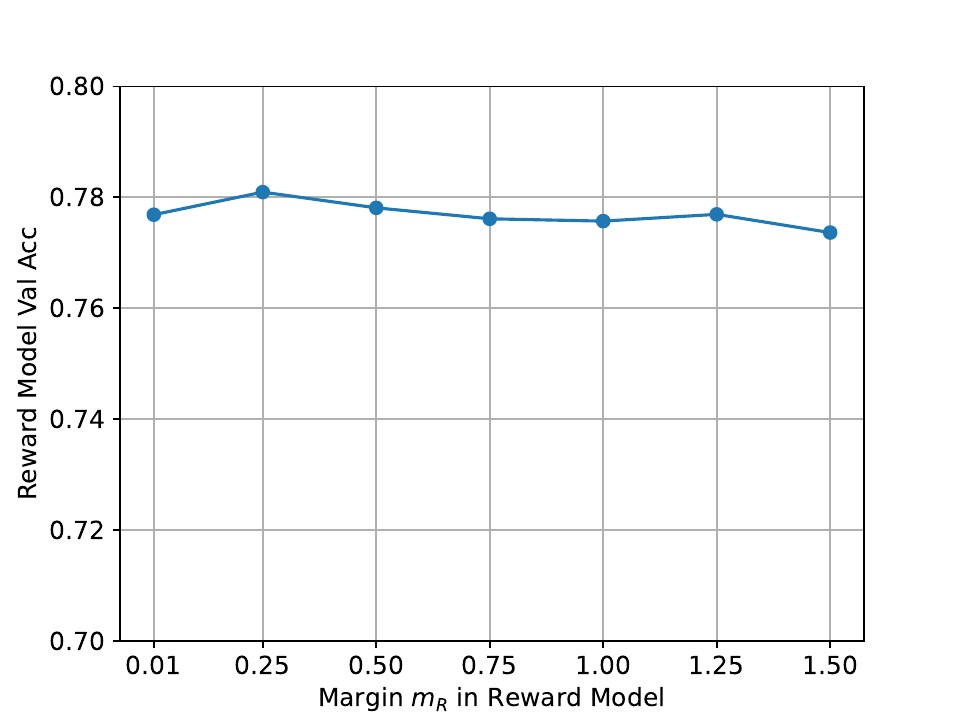}}
\subcaptionbox*{}
    {\includegraphics[width=0.3\linewidth]{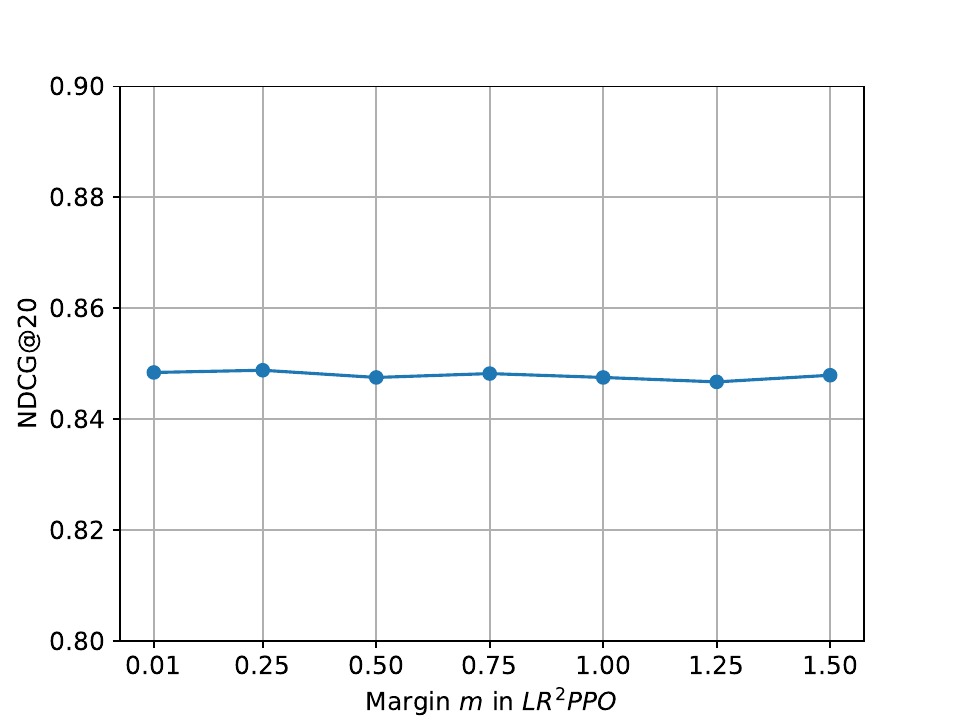}}
\subcaptionbox*{}
    {\includegraphics[width=0.3\linewidth]{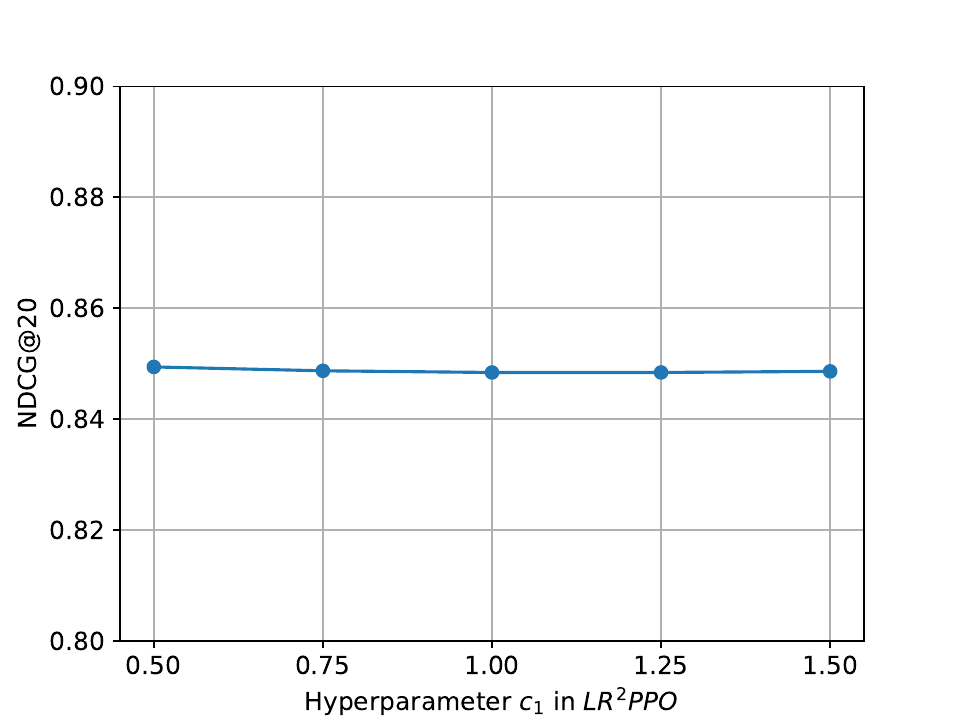}}\\
\caption{Results with different margins $m_R$, $m$, and hyperparameter $c_1$.}
\label{fig:hyperparameters}
\end{figure*}

Here, we conduct experimental analysis on the hyperparameters of the LR\textsuperscript{2}PPO algorithm, namely margin $m_R$ in reward model training, margin $m$ and hyperparameter $c_1$ in joint training of LR\textsuperscript{2}PPO.
(See Eq.~(\hyperlink{fake}{4}), Eq.~(\hyperlink{fake}{7}), and Eq.~(\hyperlink{fake}{11}) in the main paper.)
The results are shown in Fig.~\ref{fig:hyperparameters}. 
It can be observed that the final performance of LR\textsuperscript{2}PPO is not sensitive to the variations of these hyperparameters, indicating the stability of our method.

\subsection{Training Curves}

\begin{figure*}[!htb]
\centering
\subcaptionbox{Reward\label{fig:1b}}
    {\includegraphics[width=0.3\linewidth, height=0.175\textheight]{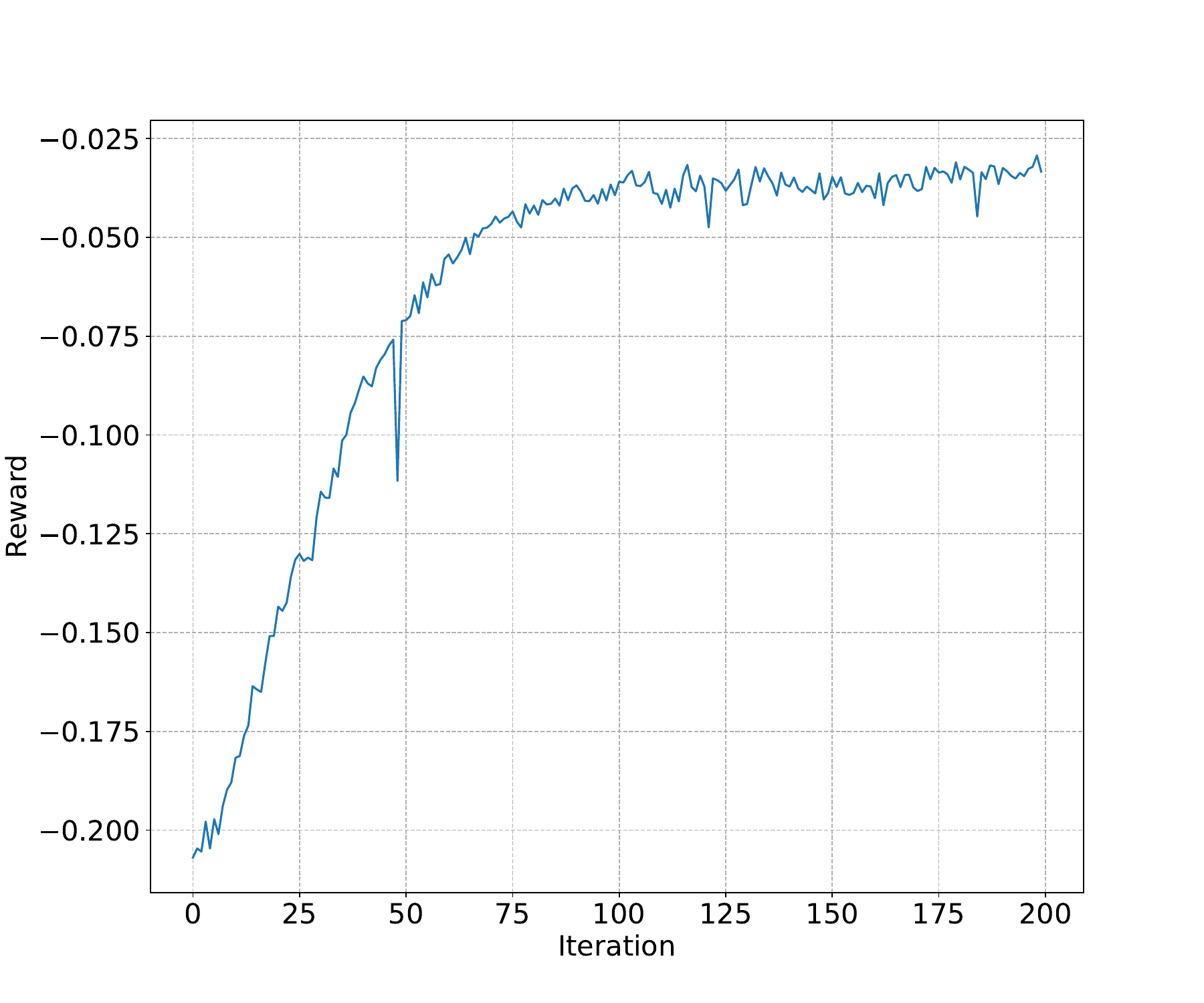}}
\subcaptionbox{Value\label{fig:2b}}
    {\includegraphics[width=0.3\linewidth, height=0.175\textheight]{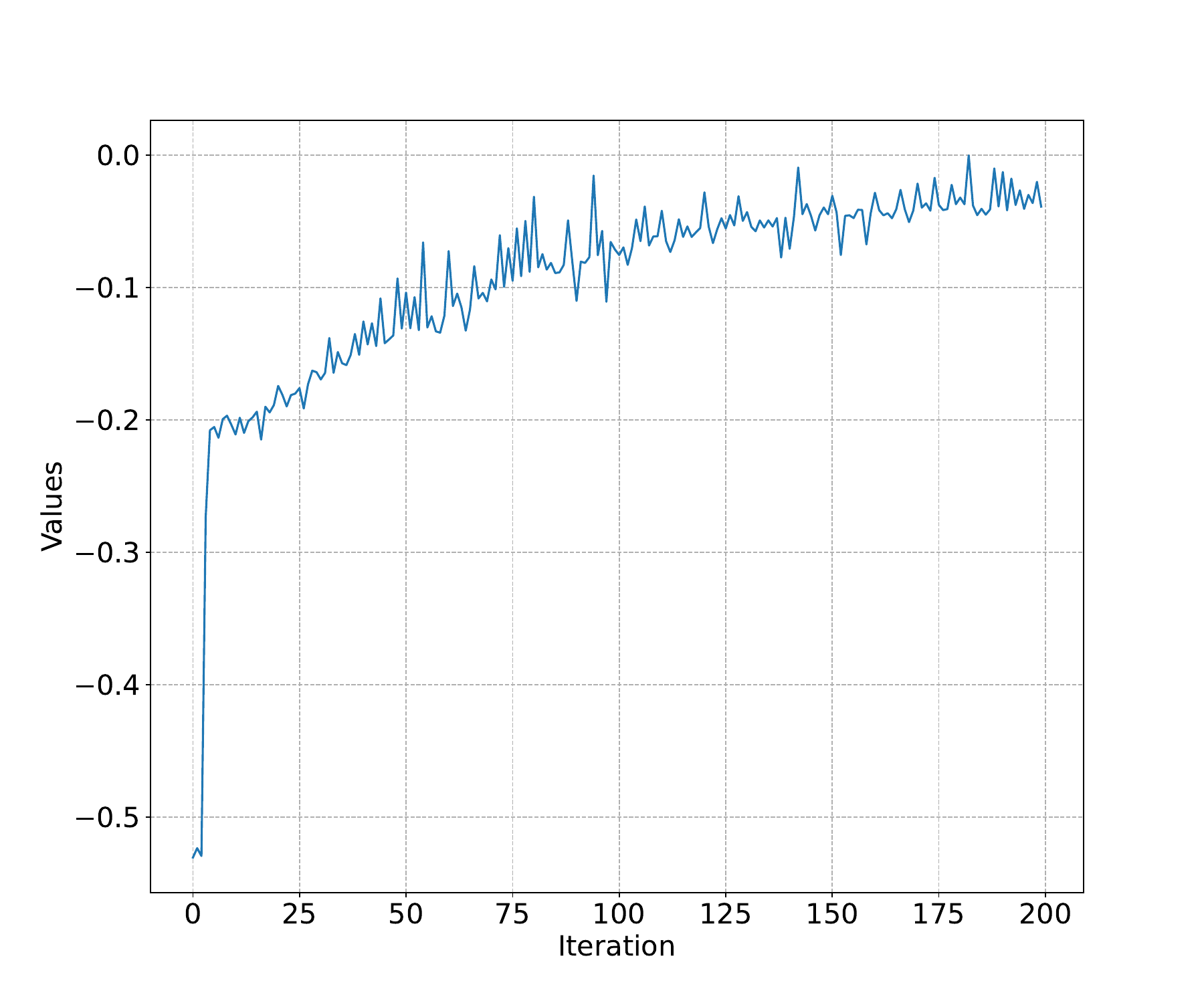}}
\subcaptionbox{Old value\label{fig:3c}}
    {\includegraphics[width=0.3\linewidth, height=0.175\textheight]{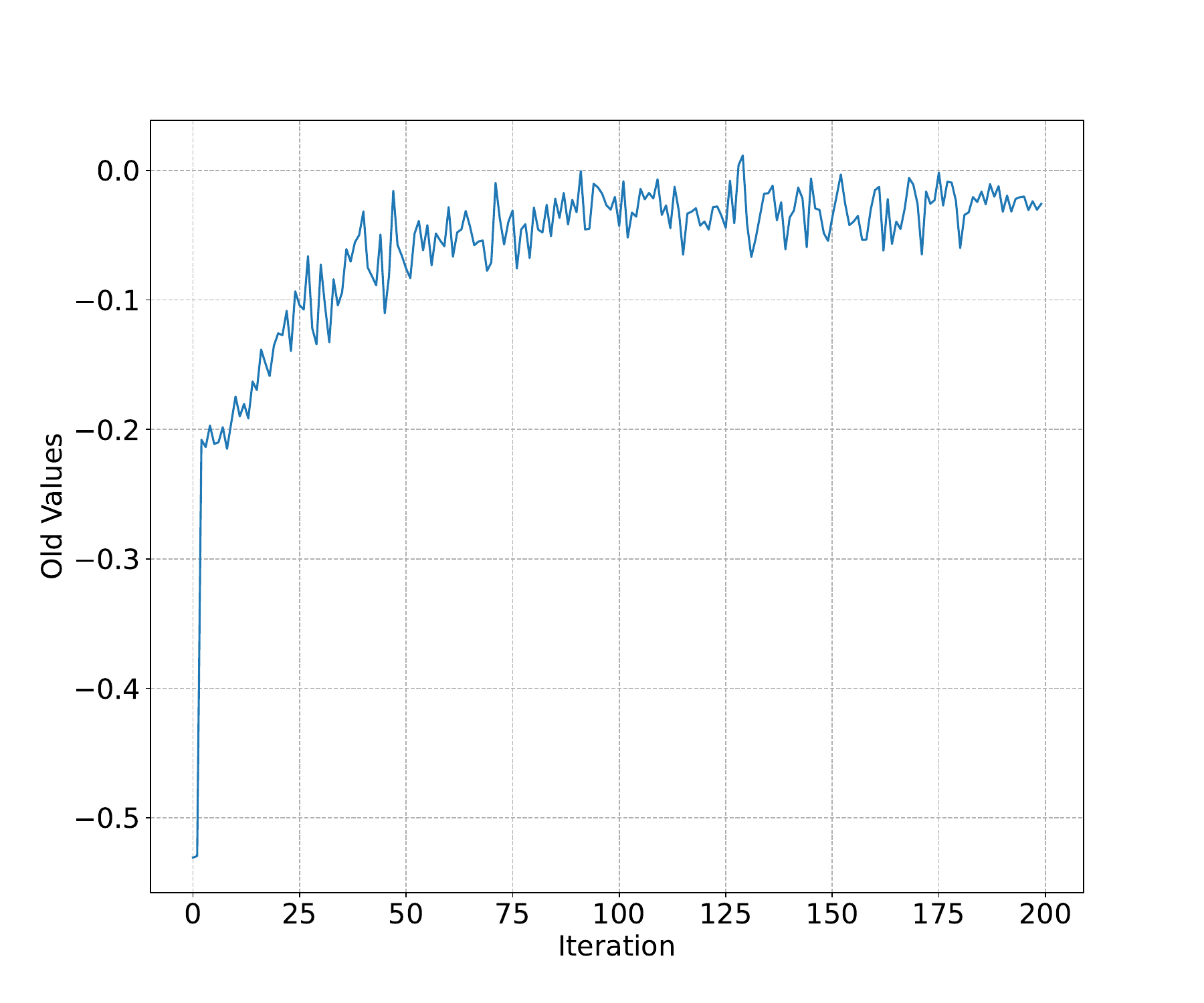}}
\\
\subcaptionbox{Advantage\label{fig:2c}}
    {\includegraphics[width=0.3\linewidth, height=0.175\textheight]{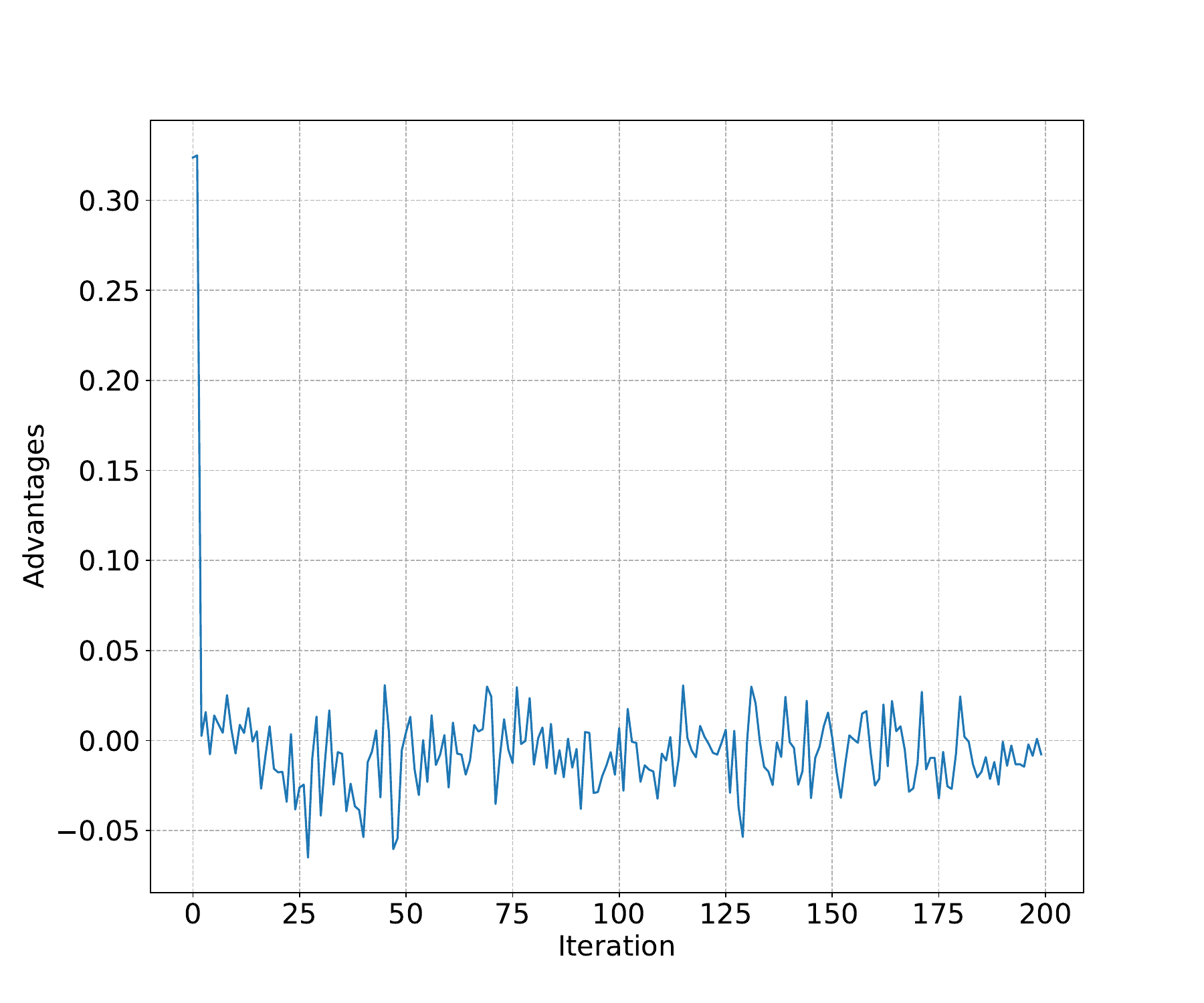}}
\subcaptionbox{Policy loss\label{fig:3a}}
    {\includegraphics[width=0.3\linewidth, height=0.175\textheight]{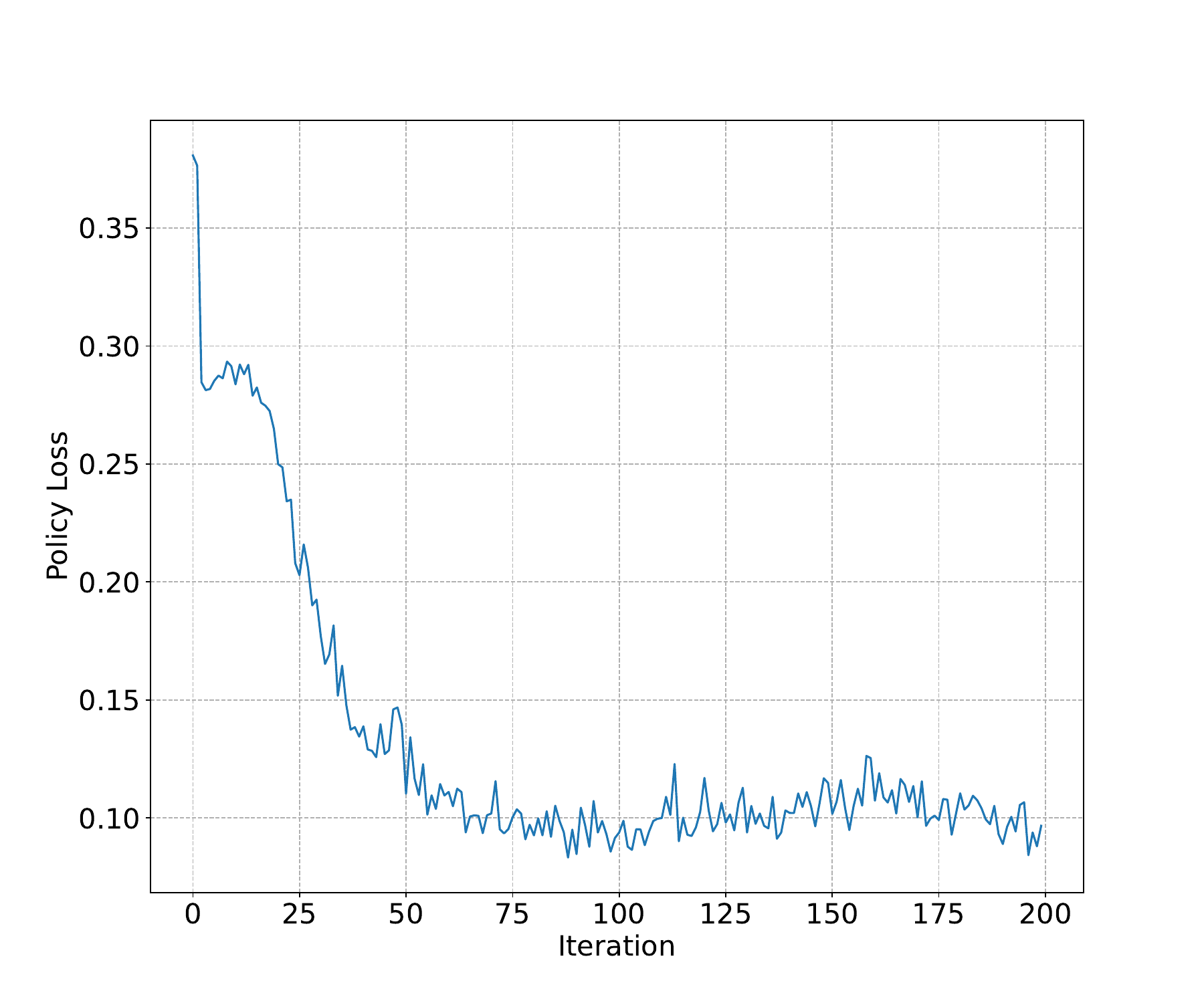}}
\subcaptionbox{Value loss\label{fig:3b}}
    {\includegraphics[width=0.3\linewidth, height=0.175\textheight]{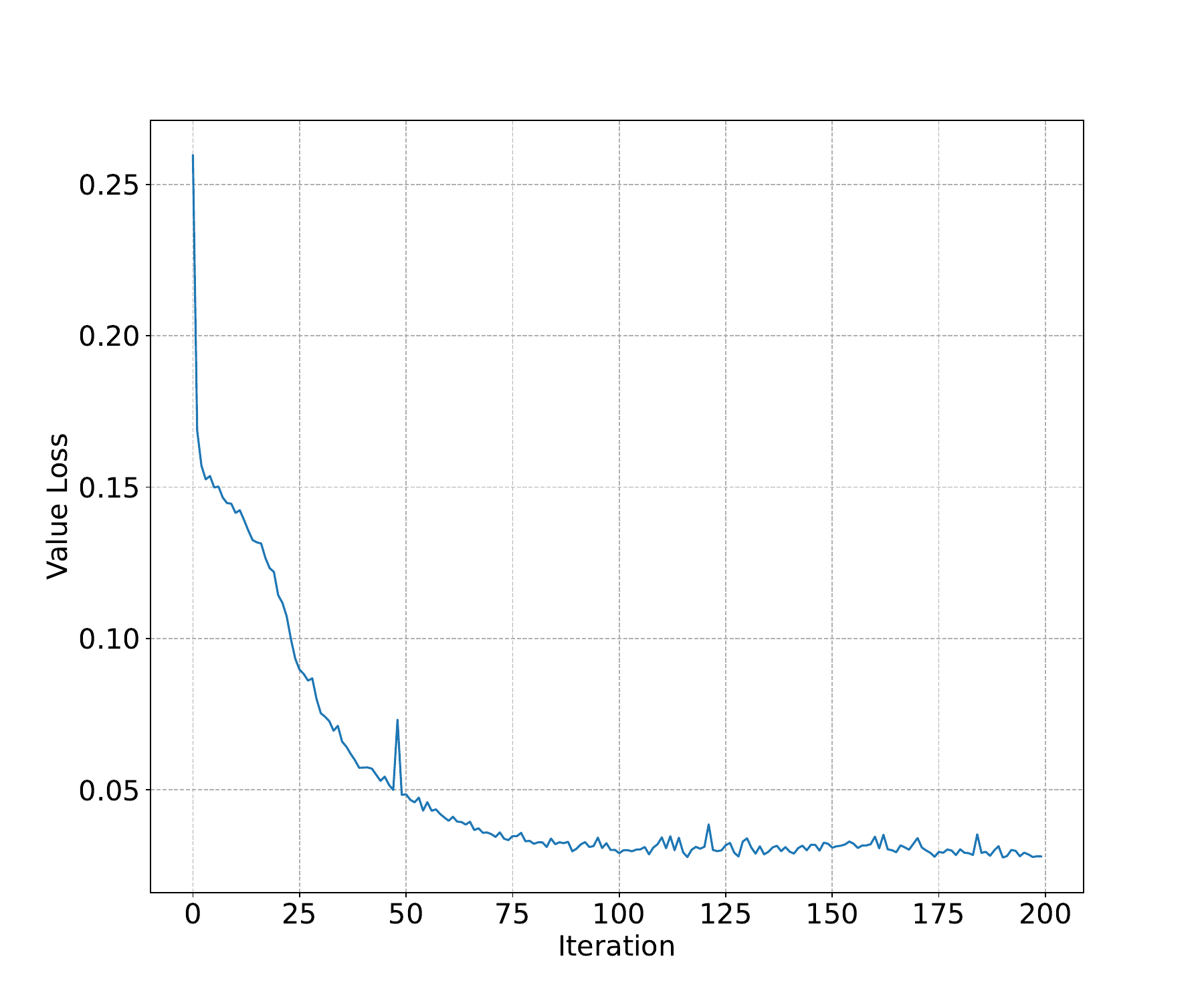}}
\\
\subcaptionbox{Partial order ratio\label{fig:4b}}
    {\includegraphics[width=0.3\linewidth, height=0.175\textheight]{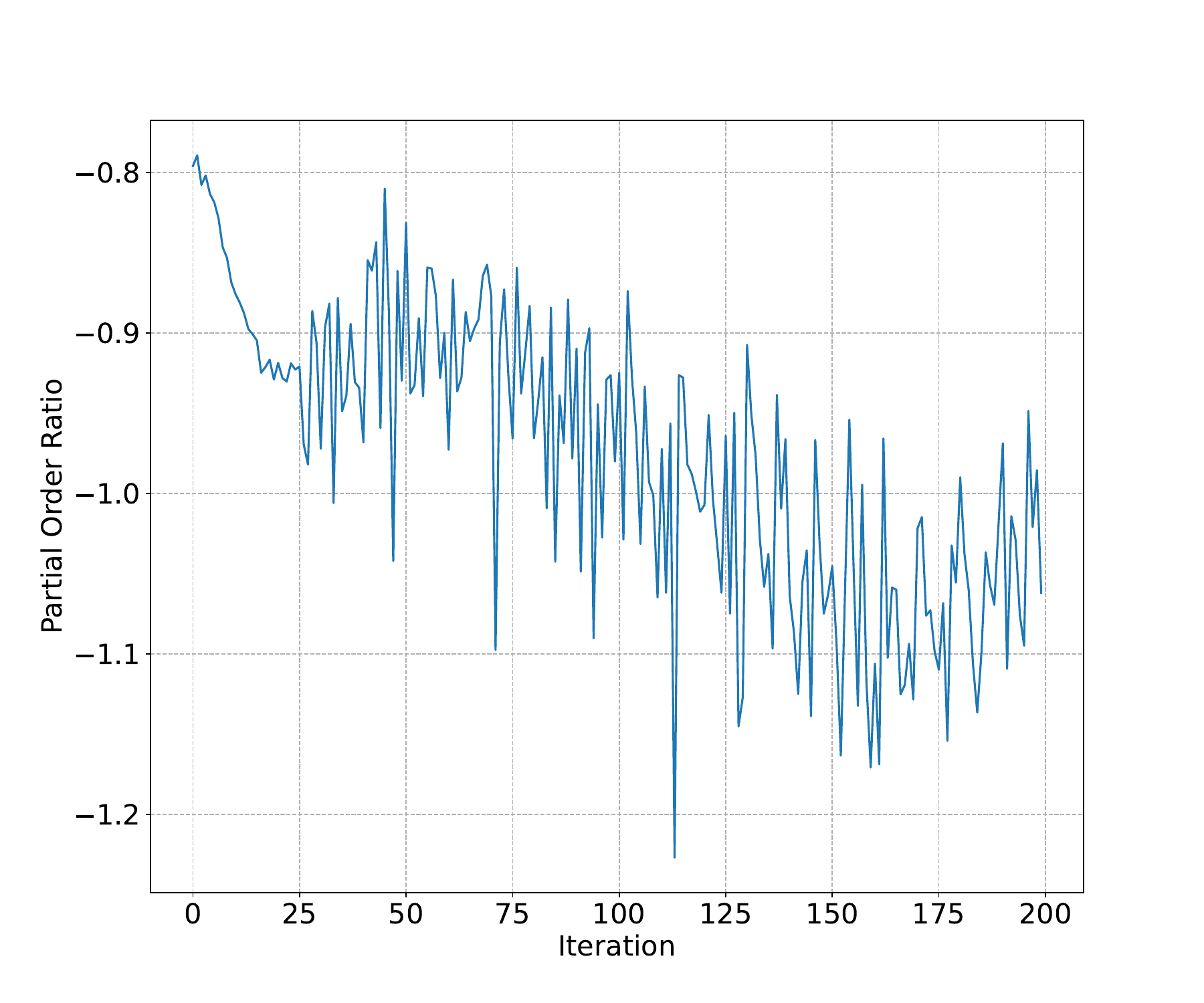}}
\subcaptionbox{Entropy bonus\label{fig:1a}}
    {\includegraphics[width=0.3\linewidth, height=0.175\textheight]{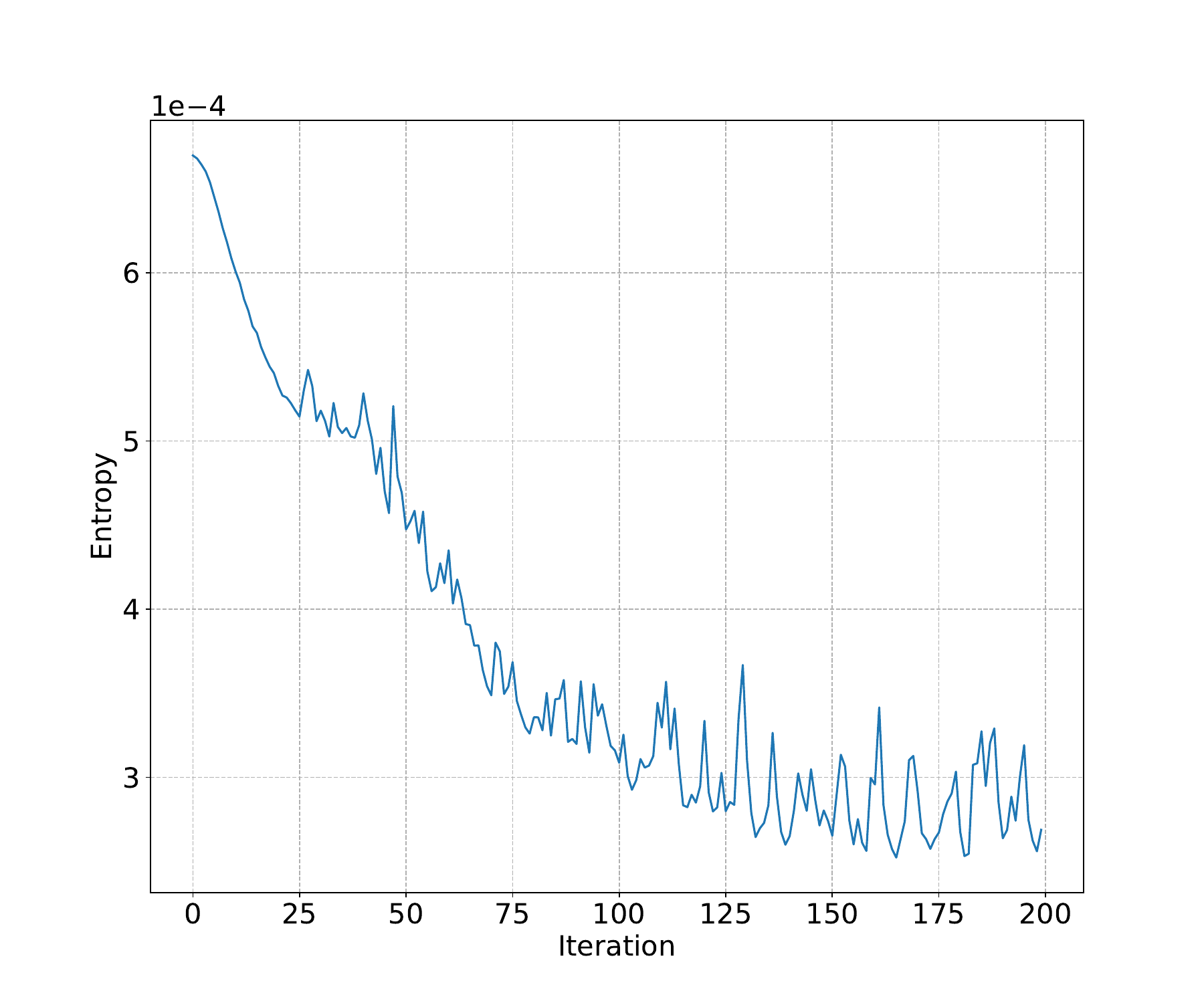}}
\subcaptionbox{KL penalty\label{fig:4a}}
    {\includegraphics[width=0.3\linewidth, height=0.175\textheight]{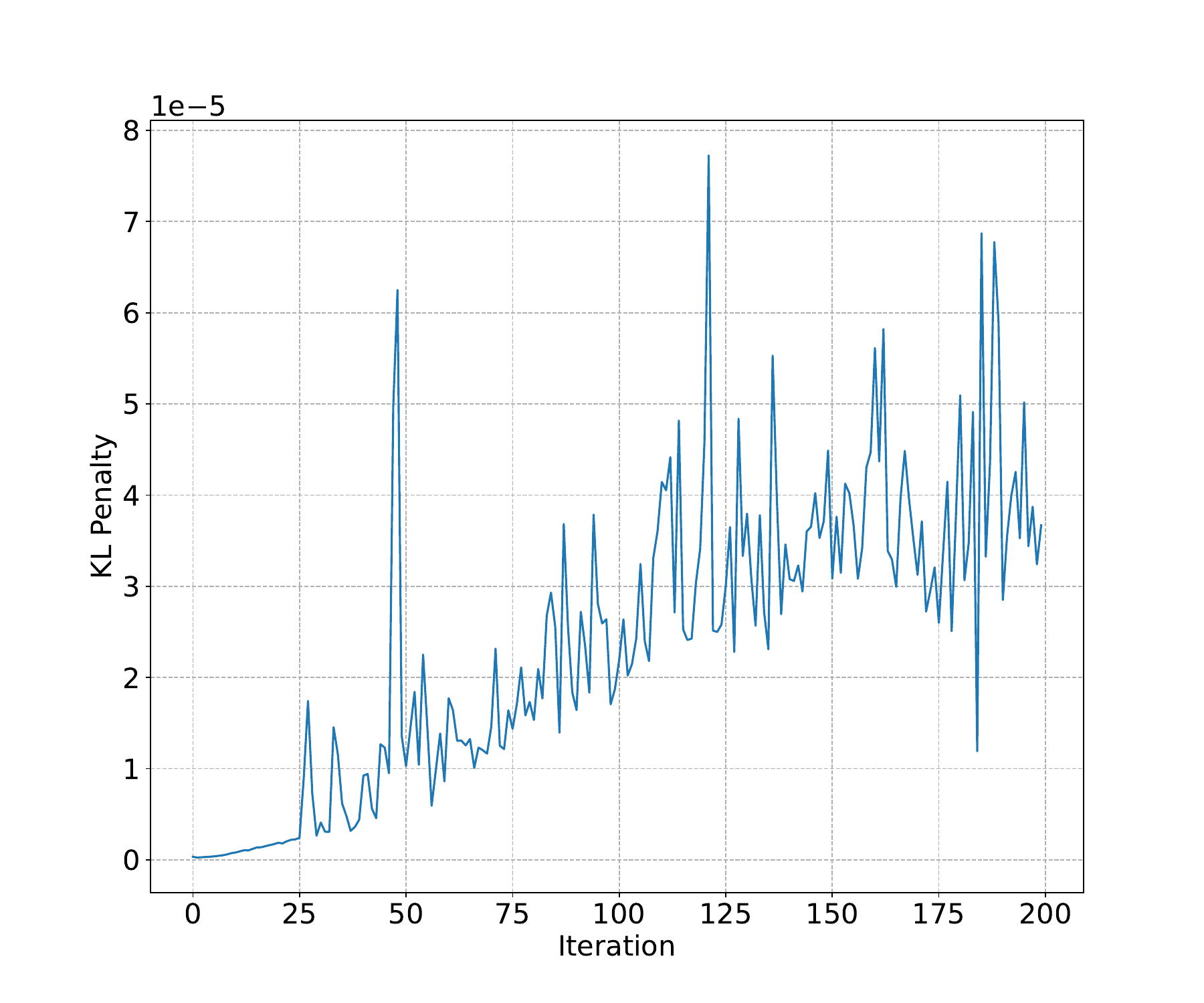}}
\\
\subcaptionbox{Reward - KL penalty\label{fig:4b}}
    {\includegraphics[width=0.3\linewidth, height=0.175\textheight]{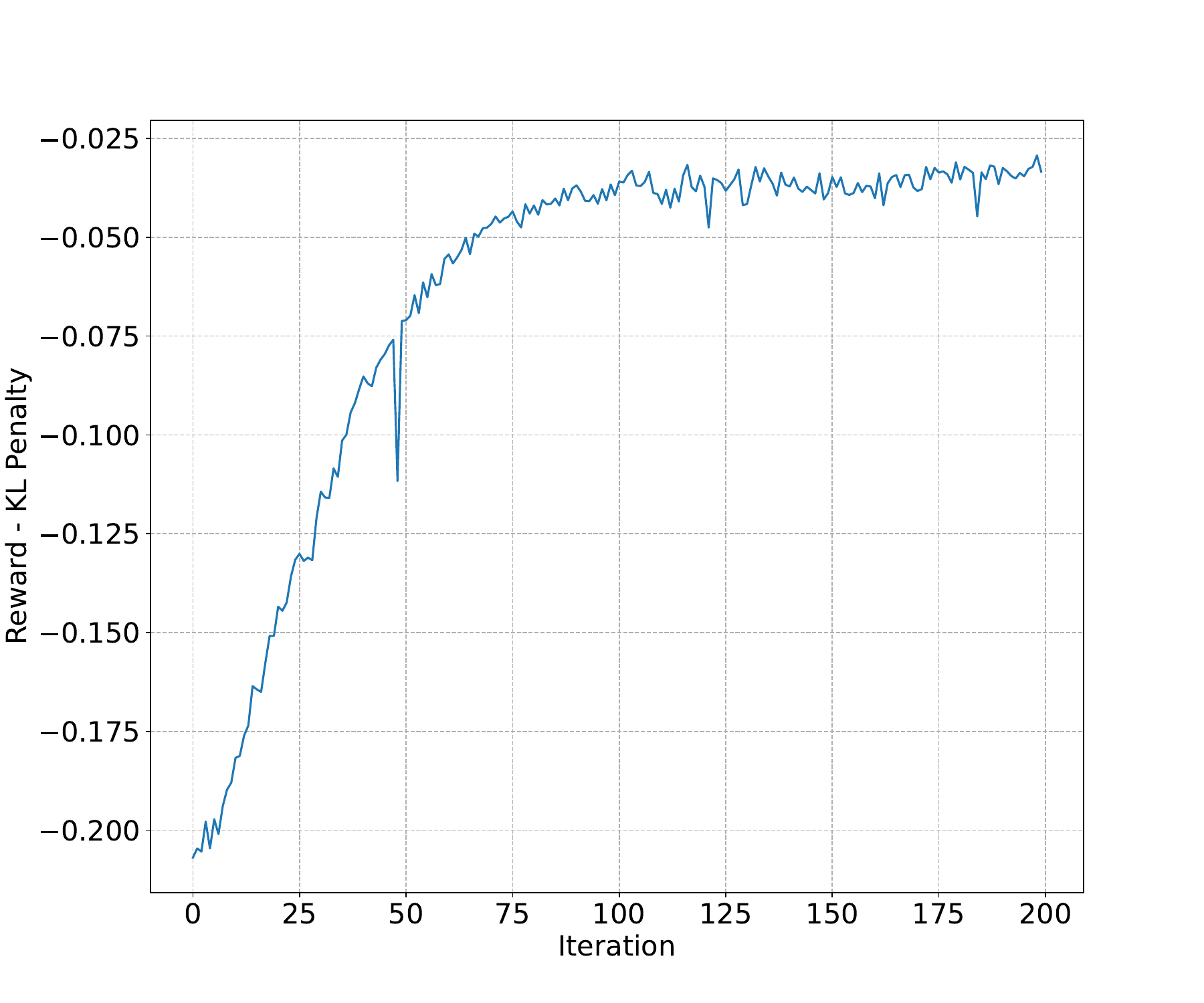}}
\subcaptionbox{NDCG@3\label{fig:4c}}
    {\includegraphics[width=0.3\linewidth, height=0.175\textheight]{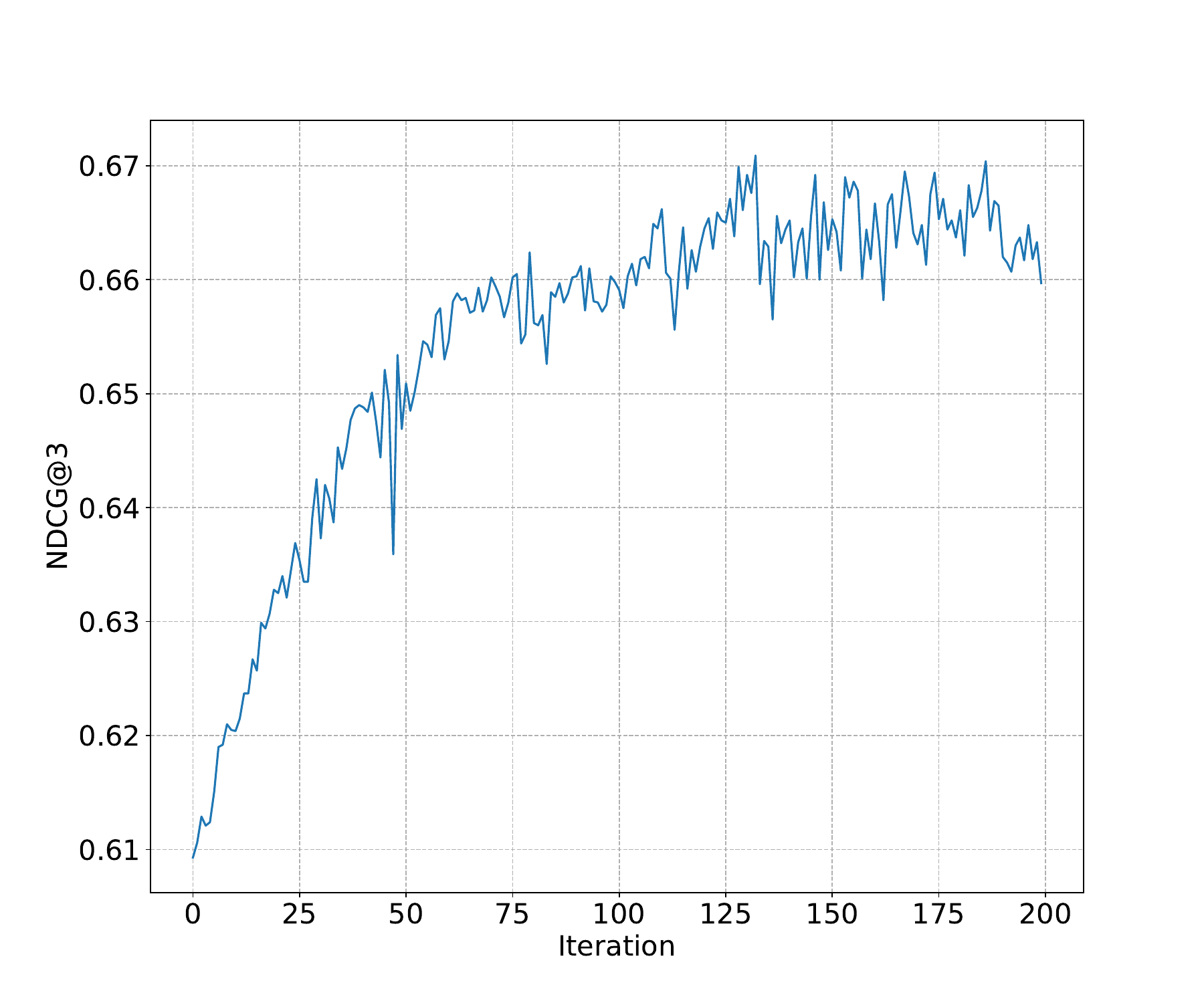}}
\subcaptionbox{NDCG@5\label{fig:5a}}
    {\includegraphics[width=0.3\linewidth, height=0.175\textheight]{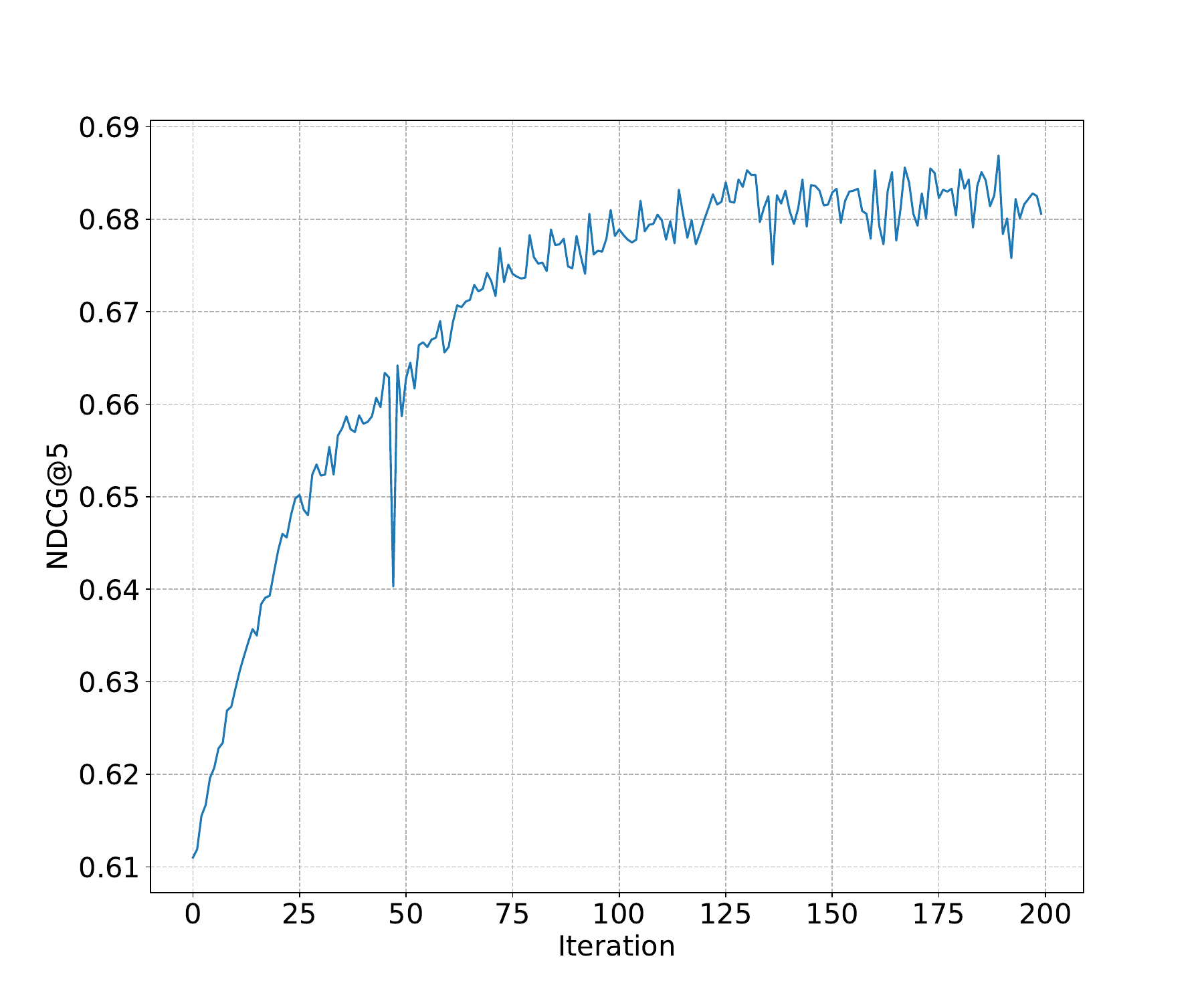}}
\\
\subcaptionbox{NDCG@20\label{fig:4c}}
    {\includegraphics[width=0.3\linewidth, height=0.175\textheight]{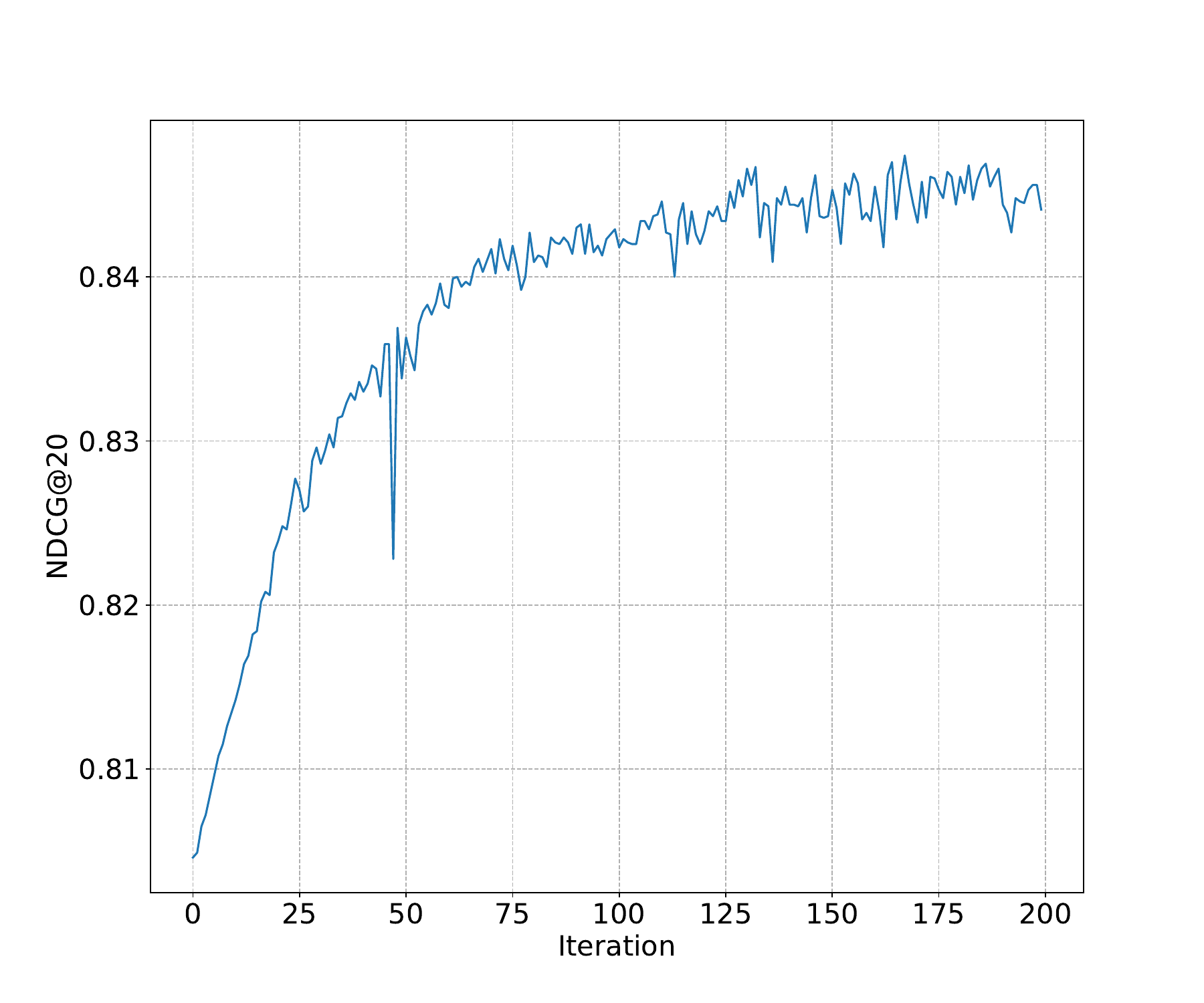}}
\subcaptionbox{Stage 2 reward accuracy\label{fig:5b}}
    {\includegraphics[width=0.3\linewidth, height=0.175\textheight]{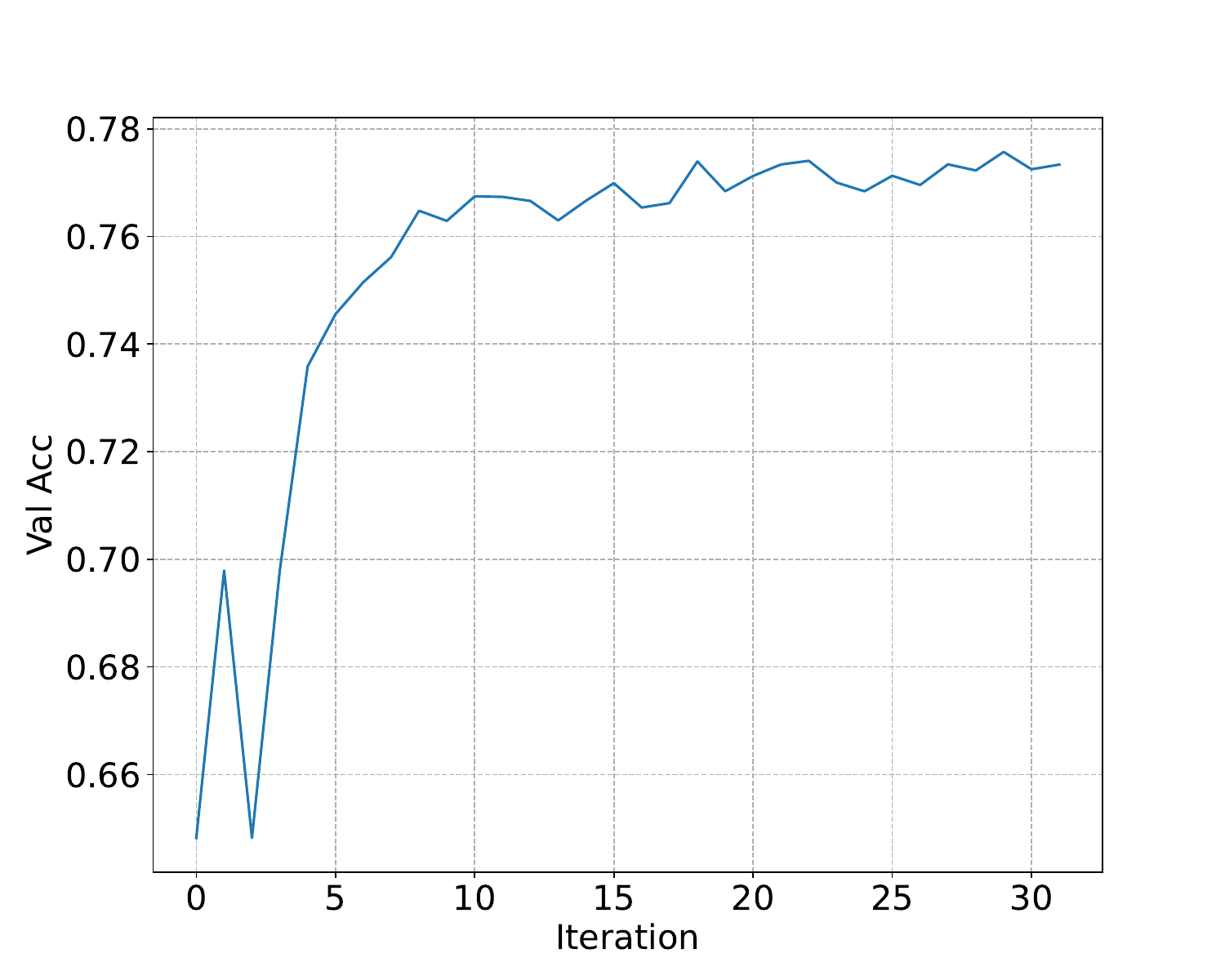}}
\subcaptionbox{Stage 1 NDCG curve\label{fig:5c}}
    {\includegraphics[width=0.3\linewidth, height=0.175\textheight]{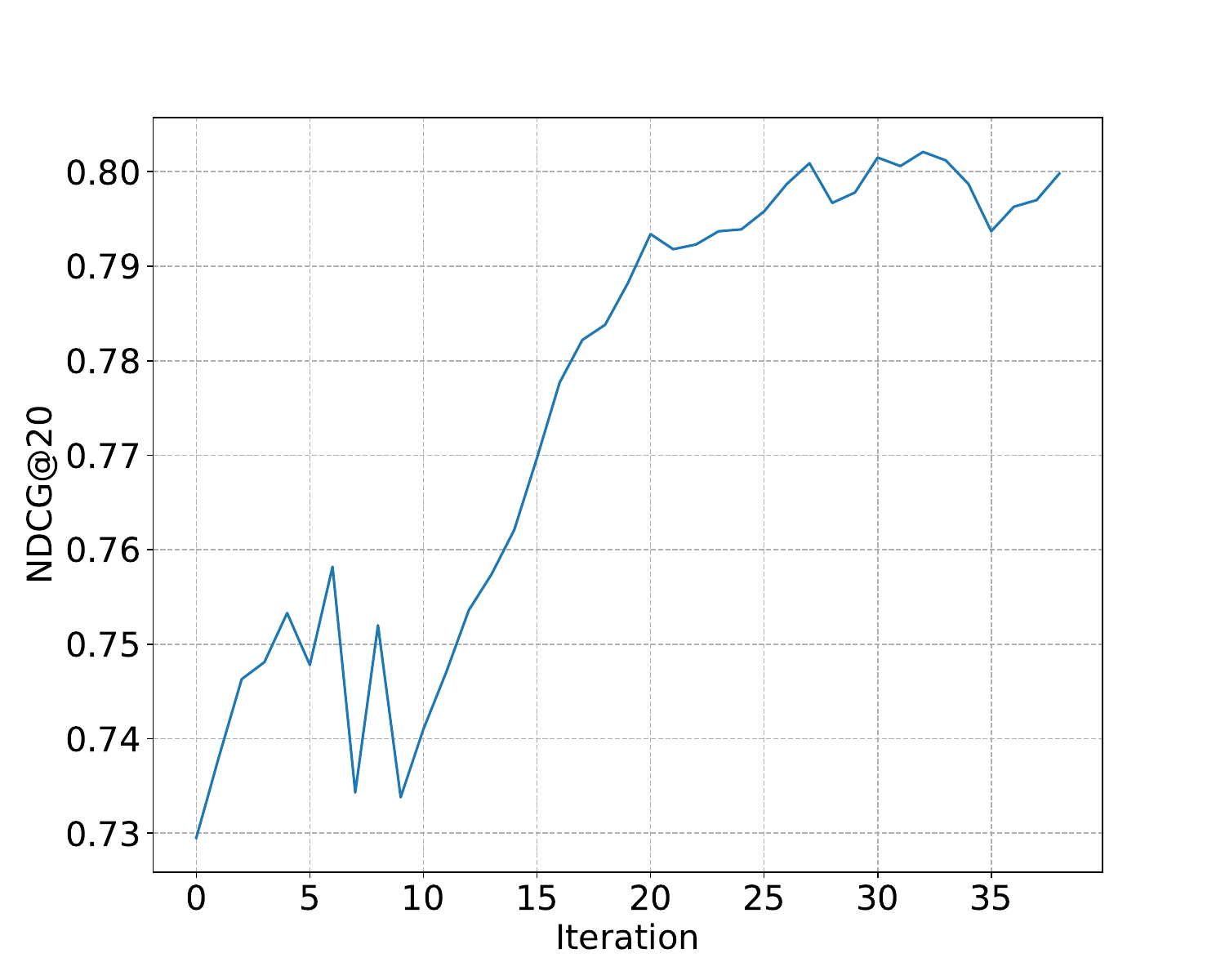}}
\caption{Training curves of LR\textsuperscript{2}PPO.}
\label{fig:training_curves}
\end{figure*}

To better illustrate the changes during the training process of the LR\textsuperscript{2}PPO algorithm, we have plotted the training curves of various variables against iterations. These are displayed in Fig.~\ref{fig:training_curves}.
Fig.~\ref{fig:1b} to Fig.~\ref{fig:4c} represent the training curves for the third stage, while \ref{fig:5b} and \ref{fig:5c} correspond to the training curves for the second and first stages, respectively.
The reward and value curves, along with the NDCG curves in the third training stage, all exhibit an upward trend, while the policy loss and value loss show a downward trend.
All of these observations validate the effectiveness of the LR\textsuperscript{2}PPO learning process.

\end{document}